\documentclass[runningheads]{llncs}

\usepackage{eccv}

\usepackage{eccvabbrv}

\usepackage{graphicx}
\usepackage{booktabs}
\usepackage{multirow}
\usepackage{pifont}
\usepackage{float}
\usepackage{wrapfig,lipsum,booktabs}
\usepackage{wrapfig}

\usepackage[accsupp]{axessibility}

\makeatletter
\def\hlinew#1{%
  \noalign{\ifnum0=`}\fi\hrule \@height #1 \futurelet
   \reserved@a\@xhline}
\makeatother

\usepackage{hyperref}

\usepackage{orcidlink}
\usepackage{colortbl}

\definecolor{gtgray}{gray}{0.97}
\definecolor{mygray}{gray}{.88}

\definecolor{gray1}{gray}{.90}
\definecolor{gray2}{gray}{.92}
\definecolor{gray3}{gray}{.94}

\makeatletter
\renewcommand*{\@fnsymbol}[1]{\ensuremath{\ifcase#1\or *\or \dagger\or \ddagger\or
   \mathsection\or \mathparagraph\or \|\or **\or \dagger\dagger
   \or \ddagger\ddagger \else\@ctrerr\fi}}
\makeatother

\begin{document}

\title{Tri$^{2}$-plane: Thinking Head Avatar via Feature Pyramid} 

\author{\thanks{Project Lead.}Luchuan Song\inst{1}\orcidlink{0000-1111-2222-3333} \and
\thanks{Equal contribution.}Pinxin Liu\inst{1} \and
Lele Chen\inst{2} \and
Guojun Yin\inst{3} \and
Chenliang Xu\inst{1}}

\authorrunning{L. Song et al.}

\institute{University of Rochester, Rochester NY 14627, USA \and
Sony AI, USA  \and
University of Science and Technology of China\\
\email{lsong11@cs.rochester.edu}}

\maketitle

\begin{center}
    \centering
    \captionsetup{type=figure}
    \vspace{-.4cm}
    \includegraphics[width=.95\linewidth]{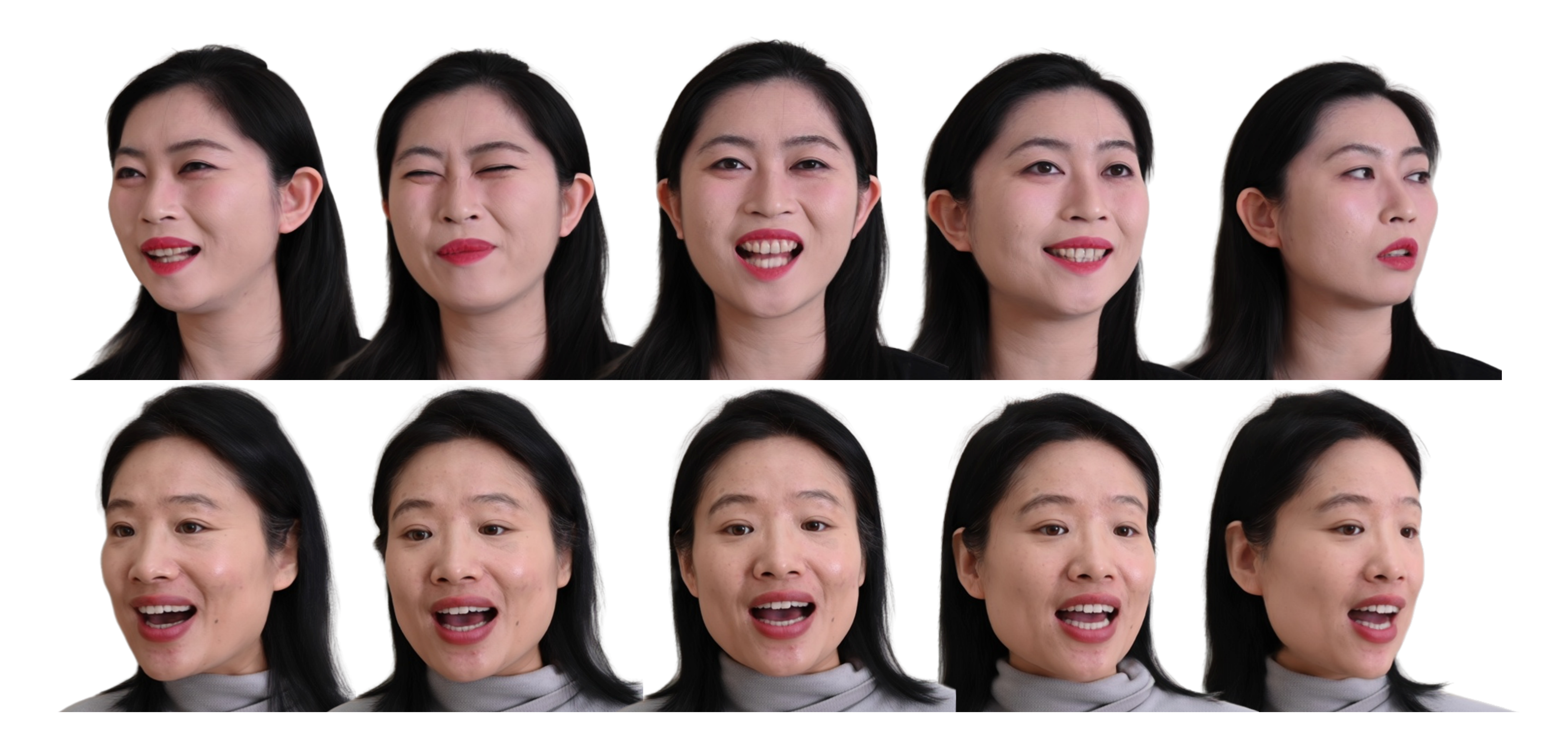}
    \vspace{-0.1cm}
    \caption{
        \textbf{We present Tri$^2$-plane}, a method designed for high-fidelity head avatar reconstruction from a short monocular video. The top row illustrates the novel view avatar synthesis (interpolation of viewpoints ranging from [-$40^{\circ}$, +$40^{\circ}$]) with facial expressions, and the bottom row displays the canonical appearance at corresponding.
    } \label{fig:teaser}
    \vspace{-.3cm}
\end{center}
\vspace{-.3cm}

\begin{abstract}
Recent years have witnessed considerable achievements in facial avatar reconstruction with neural volume rendering. Despite notable advancements, the reconstruction of complex and dynamic head movements from monocular videos still suffers from capturing and restoring fine-grained details. In this work, we propose a novel approach, named Tri$^2$-plane, for monocular photo-realistic volumetric head avatar reconstructions. Distinct from the existing works that rely on a single tri-plane deformation field for dynamic facial modeling, the proposed Tri$^2$-plane leverages the principle of feature pyramids and three top-to-down lateral connections tri-planes for details improvement. It samples and renders facial details at multiple scales, transitioning from the entire face to specific local regions and then to even more refined sub-regions. Moreover, we incorporate a camera-based geometry-aware sliding window method as an augmentation in training, which improves the robustness beyond the canonical space, with a particular improvement in cross-identity generation capabilities. Experimental outcomes indicate that the Tri$^2$-plane not only surpasses existing methodologies but also achieves superior performance across quantitative and qualitative assessments. The project website is: \url{https://songluchuan.github.io/Tri2Plane.github.io/}.
\end{abstract}

\vspace{-.9cm}
\section{Introduction}
\label{sec:intro}
\vspace{-.3cm}

Drawing inspiration from geometry-aware generative networks~\cite{shen2021closed, chan2022efficient, harkonen2020ganspace, voynov2020unsupervised, shi20223d, sun2023next3d, song2023emotional,song2024adaptive}, the triplet latent space (tri-plane)~\cite{chan2022efficient} has the potential to model the representations of the 3D structure priors. The studies in the past year~\cite{li2023efficient, xu2023latentavatar, yu2023nofa, bai2023high, li2023one, sun2022ide,song2021fsft} have substantiated tri-plane~\cite{chan2022efficient}'s capacity of capturing and modeling the facial motion representations. However, they tend to overlook the retention of high-frequency details when upsampling from low-resolution NeRF renderings to high-fidelity images. 
We attribute this problem to the upsampled features trained with low-resolution self-supervision, which inadvertently led to a paucity of fine-grained local textural details in the 3D-rendered feature maps. The super-resolution module upsamples low-resolution rendered maps over four times, which lose high-frequency details compared with directly enhancing the full-size rendered maps. This deficiency is further exacerbated when considering the inherent variability of movements captured within in-the-wild driven videos. Such movements, which reflect the diversity of camera poses and the dynamic nature of real-world interactions, amplify the model's inability to preserve details across different orientations. 

To address these challenges, we propose the Tri$^2$-plane, a new framework tailored for restoring fine-grained facial details. It incorporates three cascaded tri-planes across multiple scales of facial features. Similar to the hierarchical structure of feature pyramids~\cite{lin2017feature}, our Tri$^2$-plane synergizes various levels of tri-planes, allowing the local feature to utilize prior information from their global counterparts. Specifically, the primary tri-plane captures global image features, while the second-level tri-plane further processes primary features from one-quarter ($\frac{1}{4}$). Subsequently, the third tri-plane extends this process, culminating in the synthesis of one-sixteenth ($\frac{1}{16}$) of the face area. Each patch is combined to construct a higher-resolution rendered image via the third tri-plane. As depicted in Fig.~\ref{fig:teaser}, this method results in images with enriched facial details, as each patch performs self-supervised refinement.

To improve the Tri$^2$-plane adaptability to unseen head positions that deviate from training videos, we propose a novel geometry-aware sliding window augmentation applied to the training footage. It formulates a general expression for the sliding window that adheres to geometric constraints under arbitrary camera settings and maps correspondences to our parameterized facial models. This integration further reinforces the robustness and adaptability of our technique, ensuring that the reenacted avatars preserve textural detail and exhibit high quality across varying poses and expressions. Note that this sliding window is a general method that can be plugged into other NeRF-based methods to further improve the performance.

As shown in Fig.~\ref{fig:teaser}, our method maintains high fidelity for canonical appearance under novel viewpoints (second row) while accurately simulating facial expressions (first row). Our contributions are summarized as follows: 
\begin{itemize}
    \item We propose a multi-scale tri-plane space, utilizing a pyramid feature structure to enhance the representational capacity for modeling fine-grained facial details. 
    \item We investigate the out-of-distribution adaptation issue and present an effective geometry-aware sliding window method to improve the robustness of tri-plane generation. 
    \item Extensive experiments have established that our method outperforms the existing state-of-the-art approaches in most quantitative metrics and qualitative outcomes.
\end{itemize}
Our method provide a novel view to explore the low quality of rendering maps problem, similar to some leading methods (\eg Tri-MipRF~\cite{hu2023tri}) but different in that our method is progressive global-to-local spatial sampling, and Tri-MipRF~\cite{hu2023tri} is cone sampling along the ray.

\begin{figure*}[t]
\begin{center}
\includegraphics[width=1.0\linewidth]{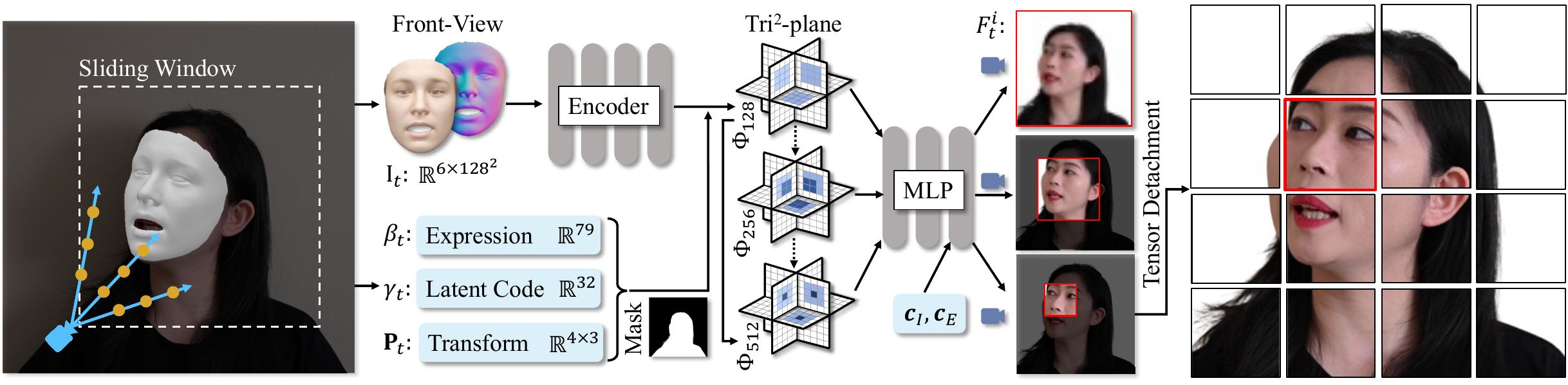}
\end{center}
\vspace{-0.5cm}
  \caption{\textbf{Overview of Tri$^2$-plane}. The pipeline steps include four components: \textbf{(1)} parametric facial tracking and zero-pose rendering are applied to generate mean texture and normal maps (shown as Front-View); \textbf{(2)} a facial condition embedding from inputs ($\beta_{t}$, $\gamma_{t}$ and encoded $\mathbf{I}_{t}$); \textbf{(3)} the multiple tri-plane for voxel rendering (as Tri$^2$-plane), accommodating various facial scales while employing shared MLP weights and \textbf{(4)} the resulting images are refined with a super-resolution model (not depicted in the figure). Furthermore, we have introduced the geometry-aware sliding window for training data augmentation to improve robustness, which incorporates the camera parameters ($\mathbf{c}_{I}$,$\mathbf{c}_{E}$) with the tracked translation values to form the training pair.
}
\vspace{-0.5cm}
\label{fig_overview}
\end{figure*}

\vspace{-0.4cm}
\section{Related Works}
\vspace{-0.3cm}

\noindent \textbf{Neural Facial Representation with Tri-plane.} The tri-plane~\cite{chan2022efficient} facilitates the extension of 2D image generation into 3D setting by integrating GANs~\cite{goodfellow2020generative} with neural radiance fields (NeRF)~\cite{mildenhall2021nerf}. The past one year has witness the tri-plane~\cite{chan2022efficient} for dynamic neural facial representation, such as Next3D~\cite{sun2023next3d}, OTAvatar~\cite{ma2023otavatar}, NOFA~\cite{yu2023nofa}, IDE-3D~\cite{sun2022ide} FENeRF~\cite{sun2022fenerf} and HFA-GP~\cite{bai2023high}~\etc. 

These methods follow standardized steps: (1) capturing the facial expression and pose through Deng et al.~\cite{deng2019accurate} from large-scale 2D datasets~\cite{karras2019style, nagrani2020voxceleb, chung2018voxceleb2}, (2) jointly training low-resolution NeRF (generally set to $64\times64$ or $128\times128$) with super-resolution module on the datasets. However, they are primarily designed for the unconditional generation of synthetic images, as StyleGAN~\cite{karras2019style, karras2020analyzing}. To achieve person-specific reconstruction, the GAN-inversion has been employed~\cite{abdal2019image2stylegan, richardson2021encoding, roich2022pivotal}. Inevitably, such adaptation results in a compromise on the quality of the generated images. Furthermore, they oversample the low-resolution rendered maps by NeRF more than four times to achieve high-quality ones, which reduces high-frequency details.

\noindent \textbf{Parametrical Facial Model.} The parametric facial model serves as an explicit facialization approach for reconstructing 3D faces from RGB images. The PCA-based representation is employed, such as BFM~\cite{blanz2023morphable}, FLAME~\cite{FLAME:SiggraphAsia2017}, and FaceVerse~\cite{wang2022faceverse}~\etc. The facial shape \(S\) is articulated as:
\begin{equation}
S = S(\alpha, \beta) = \bar{S} + B_{id}\alpha + B_{exp}\beta,
\label{3DMM_can2world}
\end{equation}
where \(\bar{S} \in \mathbb{R}^{3E}\) denotes the average shape, \(E\) representing number of vertices. \(B_{id}\) and \(B_{exp}\) are the PCA bases for identity and expression, respectively, and \(\alpha\), \(\beta\) are the corresponding parameters. The model facilitates spatial positioning adjustments through an Euler rotation matrix \(\mathbf{R} \in \mathbb{R}^{3\times3}\) and a translation vector \(\mathbf{T} \in \mathbb{R}^{3}\). The spatial coordinates \(S_{x,y,z}\) of the facial model is:
\begin{equation}
S_{x,y,z} = \mathbf{R} \cdot S(\alpha, \beta) + \mathbf{T}_{x,y,z},
\end{equation}
enabling the projection of the 3D facial model onto a 2D plane, utilizing a predefined camera within the scene.

We incorporate the geometry estimation method by Guo et al.~\cite{guo2018cnn} due to its exceptional efficiency and comprehensive coverage of facial expression variations.

\noindent \textbf{Feature Pyramids.} Feature pyramids have been pivotal in advancing modern detectors for several years. Contemporary research on feature pyramids predominantly falls into two categories: top-down or bottom-up architectures~\cite{he2017mask, lin2017focal, lin2017feature}, and attention-based methods~\cite{kong2018deep, zhang2020feature, zhao2021graphfpn, song2021tacr}. The Feature Pyramid Network (FPN)~\cite{lin2017feature}, a seminal model in this domain, introduces a top-down architecture with lateral connections, ensuring that each pyramid level incorporates high-level semantic information. Subsequent studies~\cite{pang2019libra, kong2018deep, zhang2020feature} have built upon the FPN, striving for more effective multi-scale feature fusion strategies. These methodologies effectively leverage the inherent strengths of feature pyramids in capturing a rich hierarchy of features.

Despite the prevalent application of these methods in detection and segmentation tasks, their utilization in 3D generation has remained unexplored. To our knowledge, our Tri$^2$-plane framework represents the first instance of applying feature pyramid principles to tri-plane rendering.\par

\vspace{-0.5cm}
\section{Overview}
\vspace{-0.3cm}

The Tri$^2$-plane, designed for high-fidelity person-specific avatar reconstruction from short monocular videos, consists of three steps: \textbf{(1)} Per-frame expression-related image estimation $I_t$ using 3DMM denoted as $S_{t}$, where $t$ is the timestamp; \textbf{(2)} The synthesis of volume-rendered maps leveraging multi-scale tri-plane features derived from $I_t$; \textbf{(3)} The enhancement of rendered maps to photo-realistic one with self-supervised super-resolution (SR) module. Moreover, we propose a simple geometry-aware sliding window mechanism to address data adaptation for out-of-frame positions via training data augmentation.

Details about face tracking, framework input/output processes, the Tri$^2$-plane, and the geometry-aware sliding window are provided in Sec.~\ref{sub: Data Stream}, Sec.~\ref{Tri$^2$-plane}, and Sec.~\ref{Geometry-Aware Sliding Window}, respectively. Additionally, In Sec.~\ref{Network Training}, we discuss the training and the overarching structure of our framework.

\vspace{-0.3cm}
\subsection{Data Stream}
\label{sub: Data Stream}
\vspace{-0.1cm}

\noindent \textbf{Parametric Face Tracking.} We use a 3D Morphable Model (3DMM)~\cite{blanz2023morphable, gerig2018morphable} to track the face expression parameters ($\beta_{t}$) and a transformation matrix $\mathbf{P}_{t}$ (includes a rotation matrix $\mathbf{R}$ and a translation vector $\mathbf{T}_{x,y,z}$ in Eqn.~\ref{3DMM_can2world}) of each input monocular video frame. The physical camera adopted for sampling is predefined by intrinsic ($\mathbf{c}_{I} \in \mathbb{R}^{3\times3}$) and extrinsic ($\mathbf{c}_{E} \in \mathbb{R}^{4\times3}$) parameters, which are the inherent attribution and spatial position of the camera.

The tracking pipeline renders the facial model in a front-view/zero-pose, incorporating mean texture and normal maps. These renderings are then channel-concatenated to create the expression-related motion map $\text{I}_t$, ensuring the preservation of 3D geometry and texture color while enhancing the convergence efficiency.

\noindent \textbf{Volume Rendering.} The Tri$^2$-plane rendering contains four inputs: \textbf{(1)} the concatenated maps $I_t$ designed to capture expression-related features while maintaining a neutral pose and identity; \textbf{(2)} A learnable latent code $\gamma_{t} \in \mathbb{R}^{32}$ (unique to each frame of the monocular video), introduced to correct misalignments due to the limited expressiveness of 3DMM, as discussed in NerFace~\cite{gafni2021dynamic}; \textbf{(3)} An explicit expression coefficient extracted from the tracked 3DMM to help the $\text{I}_t$ for emotion-related feature; \textbf{(4)} Transformation matrix $\mathbf{P}_t \in \mathbb{R}^{4\times3}$ which allows us to transform camera space points to points in the canonical space. It is formulated as: 
\begin{equation}
(\mathbf{c}, \sigma) = \Phi_{scale}(\text{I}_t, \gamma_{t}, \beta_{t}; \mathbf{P}_t). 
\label{our_nerf}
\end{equation}
The $\Phi_{scale}$ represents the tri-plane at various scales, specifically $\{\Phi_{128}, \Phi_{256}, \Phi_{512}\}$. The $\mathbf{c}$ is for color and $\sigma$ is for density.

The rendered output from each $\Phi_{scale}$ is denoted as $\hat{\text{I}}^{s}$ with scale of $32\times128\times128$. The $\hat{\text{I}}^{128}$ corresponds to the downscaling of the full image ($512\times 512$$\rightarrow$$128\times 128$), $\hat{\text{I}}^{256}$ is to the downscaling of $\frac{1}{4}$ of the face ($256\times 256$$\rightarrow$$128\times 128$) and $\hat{\text{I}}^{512}$ pertains to the original patch of $\frac{1}{16}$ of the image ($128\times 128$$\rightarrow$$128\times 128$).

\noindent \textbf{Super-Resolution Module.} The input of the SR module is derived from the 16-fold concatenated $\hat{\text{I}}^{512}$ with scale at $32\times512\times 512$. The output is the RGB image at an arbitrarily higher resolution, \eg, $3\times512\times512$ or $3\times1024\times 1024$.

\vspace{-0.3cm}
\subsection{Tri$^2$-plane}
\label{Tri$^2$-plane}
\vspace{-0.3cm}

The proposed Tri$^2$-plane is motivated by the pursuit of enhancing high-frequency facial details. To this end, we build on foundational work that (1) \textit{multi-scale feature map for richer information}, examed in the works related to feature pyramids~\cite{ghiasi2019fpn, lin2017feature, liu2018path}, and (2) \textit{patch extraction and representation help the restoration of the high-frequency image}, as discussed in the research on super-resolution reconstruction~\cite{dong2015image, huang2023boosting, chan2021basicvsr, chan2022basicvsr++}. Building upon these concepts, our approach advances in enabling fine-grained detail capture through a restructured feature pyramid.

Prior studies~\cite{chan2022efficient, sun2023next3d, zhao2023havatar, xu2023latentavatar, yu2023nofa,sun2022fenerf} take low-resolution rendering maps to meet GPU memory limitation; this prevent their capacity for high-frequency detail capture. In contrast, our Tri$^2$-plane innovates upon the traditional feature pyramid, the deeper layers in a pyramid yield smaller, semantically richer features, while shallower layers produce larger, more local information. We reverse this dynamic: our framework's shallower layers with smaller-sized tri-planes concentrate on capturing global structures, while deeper layers with larger-sized tri-planes are tailored to detailed, local features essential for high-frequency detail capture. This is achieved by rescaling the full image to a smaller size in shallower tri-plane layers for self-supervision and focusing on local patches with deeper tri-plane layers, employing a higher rescale ratio to preserve fine-grained details.

\noindent Our Tri$^2$-plane employs a convolutional neural network as generator, which is based on the StyleGAN~\cite{karras2020analyzing} architecture, for synthesizing the features of the multiple cascaded tri-plane representations. Different from noise-conditioned latent embedding within StyleGAN, the facial expression map $\text{I}_t$ is encoded into latent embedding and concatenated with ($\beta_{t}$, $\gamma_{t}$), which serves to generate Tri$^2$-plane through a serial of linear layers and modulated 2D convolutions. The process of generating tri-planes can be conceptualized as a hierarchical 3D super-resolution reconstruction, moving from coarse to fine. 

After obtaining the tri-planes feature, we leverage light rays at varying densities to capture multi-scale details. The coarsest plane as $\Phi_{128}$, uses $128^2$ light rays to sample a downscaled $512\times512$ image, capturing global features. The intermediate plane as $\Phi_{256}$, details a quarter of global image ($256\times256$) using the same light rays, focusing on mid-level details. The highest resolution plane as $\Phi_{512}$, maps these rays to a $\frac{1}{16}$ patch of the global image for fine-grained details, crucial for high-frequency features. It integrates with our model's inverted feature pyramid structure, ensuring a comprehensive representation across all scales. 

Our Tri$^2$-plane efficiently generates high-resolution facial features while conserving GPU memory. For the $\Phi_{512}$ features ($128\times128$), representing the $\frac{1}{16}$ image patch, we randomly select a patch for processing, detaching the rest. The resulting $512\times512$ feature map, composed of these detailed patches, is then processed through the SR module for enhancement, as shown in Fig.~\ref{fig_overview}.

\vspace{-0.4cm}
\subsection{Geometry-Aware Sliding Window}
\label{Geometry-Aware Sliding Window}
\vspace{-0.2cm}

We are inspired by VToonify~\cite{yang2022Vtoonify}, which crops images to simulate different rotations and captures two sub-frames from a single frame to mimic camera motion. However, it focuses on image-to-image translation and assumes an infinitely distant camera, neglecting varying camera perspectives. Such assumptions, though applicable for resolving jitter across frames due to temporal inconsistency in 2D, are impractical for conditional-NeRF methods~\cite{xu2023latentavatar,gafni2021dynamic,zhao2023havatar,chan2022efficient} due to their reliance on the geometry relationship in coordinate transformation.

Instead, we propose augmenting the camera perspective directly. It addresses both translation un-robustness and inconsistency issues by adjusting the camera position and orientation, particularly in 3D generation. Specifically, each query point in canonical space is transformed to camera space through the inverse of Eqn.~\ref{3DMM_can2world} as the following: 
\begin{equation}
\vspace{-0.1cm}
M_{cam} = \mathbf{R}^{-1}\cdot (M_{can} - \mathbf{T}_{x,y,z}), 
\label{inverse_canon}
\end{equation}
where the $M_{can}$ is the query points in canonical space and $M_{cam}$ is in camera space, $\mathbf{R}$ and $\mathbf{T}_{x,y,z}$ are from $\mathbf{P}_t$. The $M_{cam}$ is sampled to world space through the pre-defined camera $\mathbf{c}_{I}$, $\mathbf{c}_{E}$. For arbitrary camera, the $M_{world}$ is:
\begin{equation}
\vspace{-0.1cm}
M_{world} = \mathbf{c}_{I} \cdot M_{cam} + \mathbf{c}_{E}, 
\label{c2w}
\end{equation}
then, we will have:
\begin{equation}
\vspace{-0.1cm}
M_{world} =\mathbf{c}_{I} \cdot [\mathbf{R}^{-1}\cdot (M_{can} - \mathbf{T}_{x,y,z})] + \mathbf{c}_{E}.
\label{c2w}
\end{equation}
In the world coordinate space, we use $\{x, y, z\}$ to represent $M_{world}$, $M_{world} \Leftrightarrow M_{x,y,z}$. Constructing the sliding window setting, we move the $M_{x,y,z}$ in world space, and the movement vector is $[\Delta x, \Delta y, \Delta z]$ ($\Delta x$ and $\Delta y$ are pixel distance, $\Delta z$ is fixed to $0$ because the imaging plane of the camera is fixed). From the Eqn.~\ref{c2w}, we have:
\vspace{-0.2cm}
\begin{equation}
\begin{aligned}
[\Delta x, \Delta y, \Delta z] &= M_{x+\Delta x,y+\Delta y,z+\Delta z} - M_{x,y,z} \\
&= \mathbf{c}_{I} \cdot \mathbf{R}^{-1} \cdot (\mathbf{T}_{x, y, z} - \mathbf{T}_{x+\Delta x, y+\Delta y, z}).
\label{c2w_1}
\end{aligned}
\end{equation}
The $\mathbf{T}_{x,y,z}$$\in$$\mathbb{R}^{3}$ is defined in morphable model~\cite{blanz2023morphable} as movement of face geometry in world space, $\mathbf{T}_{x,y,z}$$=$$(T_x, T_y, T_z)$. Then, the $\mathbf{T}_{x,y,z}$ exhibits linearity: 
\begin{equation}
\vspace{-0.1cm}
\begin{aligned}
\vspace{-0.1cm}
[\Delta x, \Delta y] = \mathbf{c}_{I} \cdot \mathbf{R}^{-1} \cdot [\Delta T_{x}, \Delta T_{y}].
\label{c2w_2}
\end{aligned}
\end{equation}
The Eqn.\ref{c2w_2} is equivalent to scalar mapping as: 
\begin{equation}
\begin{aligned}
     [\Delta T_{x},\Delta T_{y}] &=[\mathbf{c}_{I} \cdot \mathbf{R}^{-1}]^{-1} [\Delta x, \Delta y] \\
     &= [\mathbf{R} \cdot \mathbf{c}_{I}^{-1}] [\Delta x, \Delta y]. 
\end{aligned}
\label{c2w_3}
\end{equation}
The pixel translation of the sliding window is equivalent to the adjustment in 3DMM coefficients. 
The Eqn.~\ref{c2w_3} represents a general adoption to any arbitrary predefined camera. In specific settings, it is equivalent to moving the center position value of the camera in $\mathbf{c}_{I}$. The sliding window augmentation enriches the training samples through the introduction of spatial variability and also extend the distribution value range of $T_x$, $T_y$, which addresses the distribution bias from head movements.

\vspace{-0.4cm}
\subsection{Network Training}
\label{Network Training}
\vspace{-0.2 cm}

\noindent \textbf{Network Structure.} There are four components in the framework. An encoder with two concatenated 2D convolution blocks is designed for facial embedding from $I_t$, three cascaded StyleGAN-based~\cite {karras2020analyzing} networks, each incorporating 2D convolution and four MLP layers, are employed to generate multi-scale tri-plane features. Then, three neural radiance fields share an MLP for density and color from each scale. Last, the super-resolution module includes several 2D convolutions with skip connection as U-Net~\cite{ronneberger2015u}. 

\noindent \textbf{RGB Loss.} The $l_1$ distance between the first three channels of 32-channel volume rendering maps and ground-truth images across different scales:
\begin{equation}
\vspace{-0.1cm}
\mathcal{L}_{rgb} = \sum_{s} ||\hat{\text{I}}^{s} -  \text{I}_{GT}^{s}||_{1},
\label{loss_rgb}
\end{equation}
where $\text{I}_{GT}^{s}$ denotes the different patches on ground-truth images, the scale of each is $3\times128\times128$ and $s$$\in$$\{128,256,512\}$.

\noindent \textbf{Perception Loss.} We employ the LPIPS loss~\cite{johnson2016perceptual} between synthesized and ground-truth images. It processes the full-face image at $512\times512$ resolution by combining $128\times128$ patches from $\Phi_{512}$ through tensor detachment, enabling backpropagation on active patches. Subsequently, the SR module processes this composite to produce detail-enhanced images $\text{I}_{sr}$, effectively reducing color inconsistencies across different patches. The $\mathcal{L}_{perp}$ is defined as the distance between $\text{I}_{SR}$ and the $\text{I}_{GT}$, full ground-truth image:
\begin{equation}
\vspace{-0.1cm}
\mathcal{L}_{perp} = \text{LPIPS}(\text{I}_{SR}, \text{I}_{GT}).
\label{loss_perp}
\end{equation}

\noindent \textbf{Mask Loss.} Each patch is associated with a corresponding ground-truth mask $\text{I}_{mask}^{512}$, then $l_2$ supervision is provided:
\begin{equation}
\vspace{-0.1cm}
\mathcal{L}_{mask} = ||\text{I}_{mask}^{512} - \hat{\text{I}}_{mask}^{512}||_2,
\label{loss_mask}
\end{equation}
where $\hat{I}_{mask}^{512}$ denotes the volume rendered masks of $512 \times 512$ resolution.
We trained the whole framework end-to-end and regulated the losses through weight balancing: 
\begin{equation}
\vspace{-0.1cm}
\mathcal{L}_{total} = \mathcal{L}_{rgb} + \lambda_{perp}\mathcal{L}_{perp} + \lambda_{mask}\mathcal{L}_{mask}.
\label{loss_reg}
\end{equation}
The $\mathcal{L}_{rgb}$ and $\mathcal{L}_{perp}$ are the primary focus, with a higher weight ($\lambda_{perp} = 1$) for the latter. The mask is given lower priorities with weights $\lambda_{mask} = 0.1$.

\vspace{-0.4cm}
\section{Experiments}
\label{Experiments}
\vspace{-0.3cm}

\noindent \textbf{Implementation details.} We train the network for 800,000 iterations for a specific portrait monocular video. The Adam optimizer~\cite{kingma2014adam} is adopted for all learnable parameters with a learning rate of $1e^{-4}$. Then, we sample $64$ points along each ray, and the channel of feature maps from each $\Phi_{scale}$ is set to $32$. The sliding window augmentation is set on each iteration step with the target scale of $512\times512$ (it is randomly walked in the region of $736\times736$).

\noindent \textbf{Monocular Training Data.} Our method uses short monocular RGB video sequences. We capture 10 human subjects with a Nikon Z7 mirrorless camera at a resolution of $2560 \times 1600$ pixel with a framerate of 60 frames per second. The images are cropped to $1600\times1600$ and scaled to $736\times736$ (the center cropped $512\times512$ region is the sliding window area). The sequences have a length of about 2 min (7200 frames). We hold out the last 20\% frames to serve as a test sequence for each reconstruction. The subjects were asked to engage in normal conversation, including expressions like smiling as well as head rotations. Moreover, we select 10 different subjects which are provided by the previous tasks~\cite{Gao2022nerfblendshape,zheng2022avatar,wang2023styleavatar} and public dataset NerSemble\footnote{The NerSemble~\cite{kirschstein2023nersemble} is a multi-view dataset, we only take the front view camera (Index: $222200037$) and sentence speaking squence for experiments.}\cite{kirschstein2023nersemble} for fair comparison.

\noindent\textbf{Baselines.} We compare our method with the state-of-the-art monocular avatar reconstruction methods, including:
\begin{figure}[H]
    \centering
    \includegraphics[width=.99\linewidth]{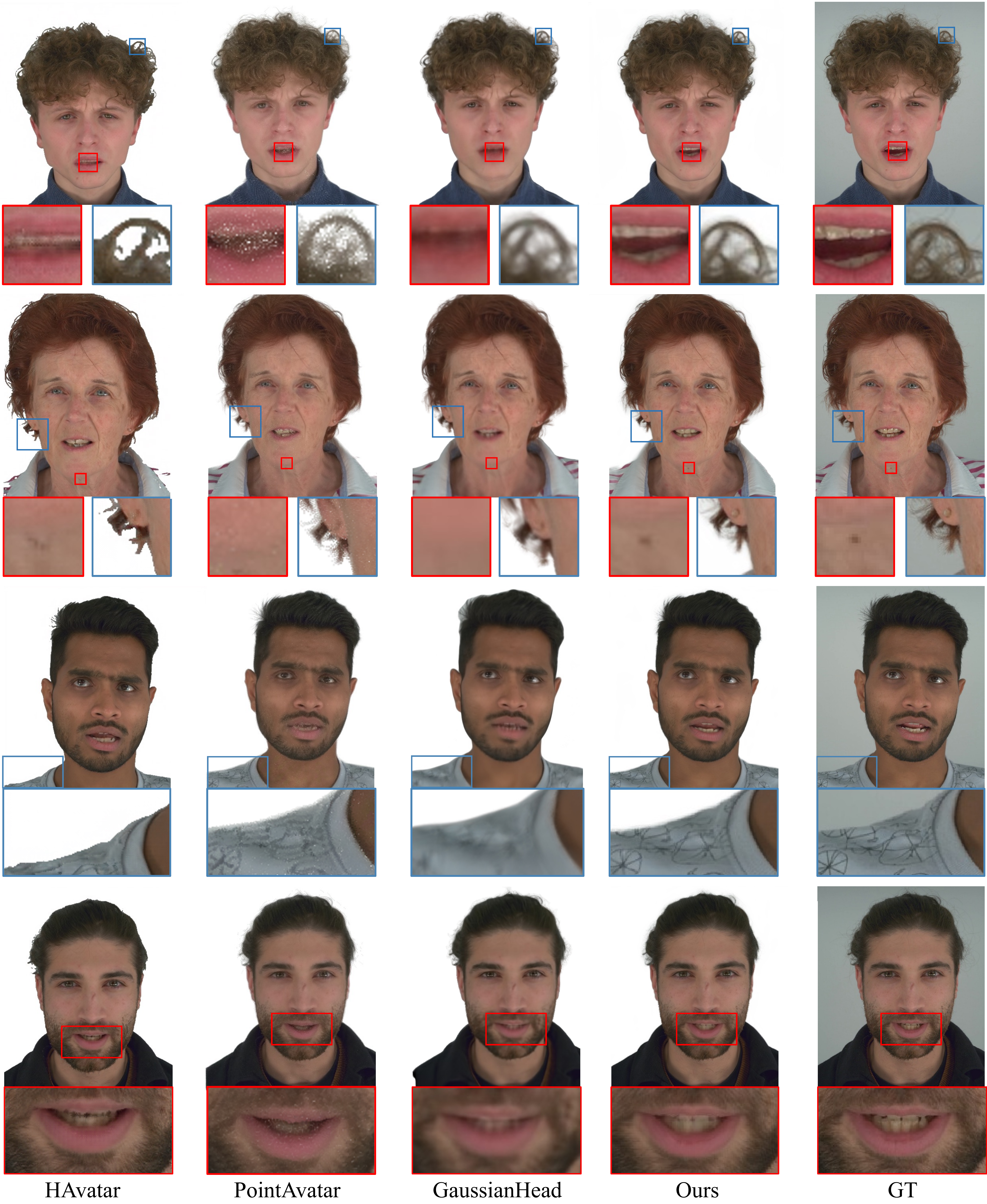}
    \vspace{-0.2cm}
    \caption{
        Qualitative comparison of different methods on the front view of the videos from NeRSemble~\cite{kirschstein2023nersemble} under self-reenactment task. From left to right: HAvatar~\cite{zhao2023havatar}, PointAvatar~\cite{Zheng_2023_CVPR}, GaussianHead~\cite{wang2024gaussianhead} and Ours. Our method achieves high-quality reconstruction details such as hair and torso textures. Please zoom in for details.
    } \label{fig:self-compare}
    \vspace{-0.9cm}
\end{figure}
\noindent \textbf{1) HAvatar}~\cite{zhao2023havatar}: The HAvatar first pretrains the single coarse triplane, and then jointly trains the SR module via an adversarial manner. We take the monocular setting in the HAvatar. 

\noindent \textbf{2) PointAvatar}~\cite{wang2023styleavatar}: The PointAvatar relies on the condition of the deformation field with head poses and expression. The coarse-to-fine strategy is to progressively increase the size of the point clouds.

\begin{figure}[H]
    \vspace{-.2cm}
    \centering
    \includegraphics[width=.9\linewidth]{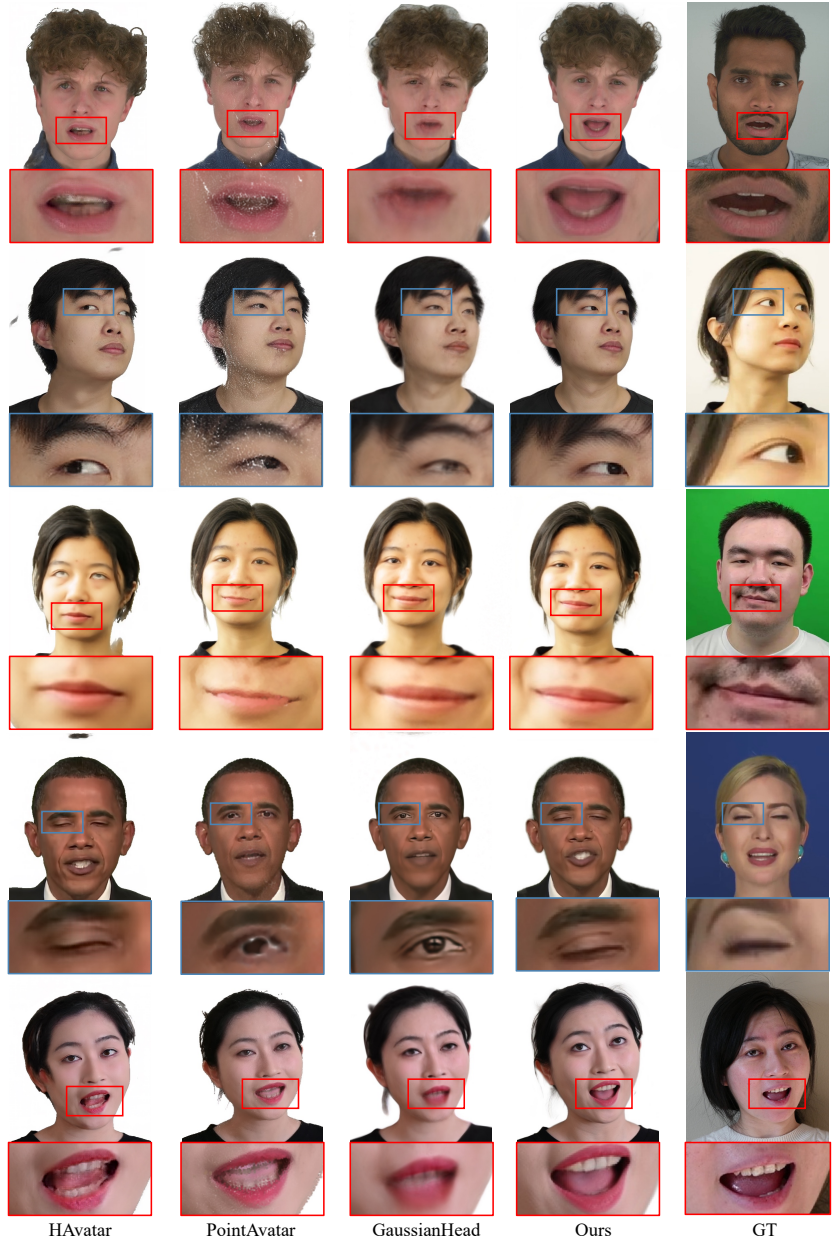}
    \vspace{-0.2cm}
    \caption{
        Qualitative comparison on the cross-reenactment. The subjects are collected from (top-down): NeRSemble~\cite{kirschstein2023nersemble}, HAvatar~\cite{zhao2023havatar}, PointAvatar~\cite{Zheng_2023_CVPR}, GaussianHead~\cite{wang2024gaussianhead} and self-recorded video. GT is the corresponding frame in the actor video. We take the official weights (HAvatar~\cite{zhao2023havatar}, PointAvatar~\cite{Zheng_2023_CVPR}) for the subjects in $2^{nd}$ and $3^{rd}$ rows. 
    } \label{fig:cross-compare}
    \vspace{-0.9cm}
\end{figure}

\noindent \textbf{3) GaussianHead}~\cite{wang2024gaussianhead}: The GaussianHead applies the 3D gaussian splatting for full head avatar reconstruction. The Gaussian points are deformed with the facial expression and head pose as the condition.

\noindent Here we have two statements about baseline methods: (1) The task is on a monocular setting, the baselines do not include reconstruction methods under multi-view video inputs and FFHQ-pretraining settings, \textit{e.g.} Gaussian-Avatar~\cite{qian2023gaussianavatars}, HQ3DAvatar~\cite{teotia2023hq3davatar}, Codec Avatars~\cite{saito2024rgca}, Next3D~\cite{sun2023next3d} and LP3D~\cite{trevithick2023} \etc, (2) We have included more methods in the Appendix and Supp. videos due to main page limitation (\textit{e.g.} \textbf{INSTA}~\cite{zielonka2023instant}, \textbf{StyleAvatar}~\cite{wang2023styleavatar}, \textbf{Deep Video Portrait}~\cite{kim2018deep}, \textbf{IMAvatar}~\cite{zheng2022avatar}, \textbf{ER-NeRF}~\cite{li2023efficient} and \textbf{RAD-NeRF}~\cite{tang2022real}), which ranges from 2D-based and deformable-NeRF based methods.

\vspace{-0.1cm}
\subsection{Quantitative Evaluation} 
\label{Quantitative Results}
\vspace{-0.1cm}

\noindent \textbf{Evaluation Metrics.} We evaluate the effectiveness of our method on three aspects: (1) F-LMD~\cite{chen2018lip}: The distance on the whole face to measure the differences in head pose and facial expression via MediaPipe~\cite{lugaresi2019mediapipe}. (2) The Sharpness Difference (SD)~\cite{mathieu2015deep}: It is used to evaluate the sharpness difference between the source and generated images, which is implemented by the pixel-level difference in rows and columns. We anticipate that a pair with good SD can exhibit equivalent changes in gradient (\textit{e.g.} shadows). (3) Image Spatial Quality: we adopt the PSNR to measure the overall image quality, the Learned Perceptual Image Patch Similarity (LPIPS)~\cite{zhang2018unreasonable} for the details. %

\noindent \textbf{Evaluation Settings.} We perform self-reenactment experiments on the self-record videos (\textbf{Dataset A}), the videos provided by previous methods~\cite{Zheng_2023_CVPR,wang2024gaussianhead,zhao2023havatar,Gao2022nerfblendshape,zheng2022avatar,wang2023styleavatar} (\textbf{Dataset B}) and NeRSemble~\cite{kirschstein2023nersemble} to quantitatively evaluate our method with the baselines. And the videos are resized to $512\times512$. We take the last 20\% frames as a holdout sequence for evaluation and other 80\% frames for training. 

\noindent \textbf{Evaluation Results.} The quantitative results are summarized in Table~\ref{table_1}. According to the results, our method outperforms the others in terms of the metrics on image quality (PSNR, LPIPS), sharpness difference (SD) and motion accuracy (F-LMD). Although there is a slight decrease in PSNR compared to HAvatar~\cite{zhao2023havatar} ($31.72$ to $31.17$), our method improves the level of detail intensity and improves perceptual similarity within local patches, as indicated by the noticeable outperformance in LPIPS and SD between ours and the baseline methods. Note that the metrics values on the Dataset A are lower than Dataset B and NeRSemble, which is because Dataset A includes more large-scale and complex head movements compared to the others. More analysis and evaluation on other six baseline methods with video visualization results are placed in Appx..

\begin{table}[t]
\footnotesize
\begin{center}
\setlength{\tabcolsep}{1.4mm}
{
\vspace{-.1cm}
\begin{tabular}{ccccccccc}
\hlinew{1.15pt}
\multirow{3}{*}{Methods} &F-LMD$\downarrow$ &SD$\downarrow$ &PSNR$\uparrow$&LPIPS$\downarrow$ &MOS$_1$ &MOS$_2$ &MOS$_3$ &MOS$_4$\\
\cline{2-9}
&\multicolumn{4}{c|}{Quantitative Results} &\multicolumn{4}{c}{User Study}\\
\cline{2-9}
&\multicolumn{4}{c|}{Dataset A + Dataset B~\cite{Zheng_2023_CVPR,wang2024gaussianhead,zhao2023havatar}}&\multicolumn{4}{c}{Self-Reenactment}\\
\hline
HAvatar & 2.94 & 3.85  & \textbf{28.72}  & \underline{8.42} & \textbf{4.49} & 4.08  & 4.21   & 4.03  \\
PointAvatar &2.70 & 7.01 & 25.85  &   15.6 & 3.52 & \underline{4.17} & \underline{4.33} &  3.94    \\
GaussianHead &\underline{2.57} & \underline{3.64} & 26.35  &  8.76 & 4.29 & 4.12 & 3.90  &  \underline{4.05}  \\
\rowcolor{mygray} Ours &\textbf{2.48} &\textbf{3.50} &\underline{27.75} & \textbf{5.81}& \underline{4.43} & \textbf{4.26} & \textbf{4.52} & \textbf{4.19} \\
\hline
&\multicolumn{4}{c|}{NeRSemble~\cite{kirschstein2023nersemble}}& \multicolumn{4}{c}{Cross-Reenactment}  \\
\hline
HAvatar & 2.30 & \underline{3.51}  & 27.32  & \underline{5.45} & 3.71 & 3.32  & 4.13  & 3.45  \\
PointAvatar &\underline{1.95} & 9.17 & 24.19 &  16.4 & 3.30 & 3.12 & \underline{4.01}  &   3.11 \\
GaussianHead & 2.17 & 4.10 & \underline{29.12}   & 5.36 &\underline{3.88} & \underline{3.42} & 3.63  &  \underline{3.50} \\
\rowcolor{mygray} Ours &\textbf{1.59} &\textbf{2.60} & \textbf{30.04}   & \textbf{4.82} &\textbf{4.01} &\textbf{3.75} &\textbf{4.22} & \textbf{3.94}  \\
\hlinew{1.15pt}
\end{tabular}}
\vspace{.3cm}
\caption{(1) Left: Quantitative results of HAvatar~\cite{zhao2023havatar}, PointAvatar~\cite{Zheng_2023_CVPR}, GaussianHead~\cite{wang2024gaussianhead} on three datasets from different sources. We bold the best and the second is highlighted with underline. Please refer to the Appx. for results with six more methods~\cite{zielonka2023instant,wang2023styleavatar,kim2018deep,zheng2022avatar,li2023efficient,tan2020efficientdet}. The experiments are performed under self-reconstruction setting. The value of SD and LPIPS are multiplied by $10^{-1}$ and $10^{2}$ respectively. (2) Right: The MOS score for human evaluation. Each one comes from a 5-point Likert scale (\textit{1-Bad, 2-Poor, 3-Fair, 4-Good, 5-Excellent}). The closer to $5$ the better, we bold the best. The experiments are performed on Dataset A and B.}\label{table_1}
\end{center}
\vspace{-1.cm}
\end{table}

We select the visualization of four actors within the NeRSemble~\cite{kirschstein2023nersemble} in Fig.~\ref{fig:self-compare}, with the corresponding area zoomed in for details. It is easy to find from Fig.~\ref{fig:self-compare} that our method
outperforms HAvatar~\cite{zhao2023havatar} in both image quality and control of facial attributes. Although HAvatar~\cite{zhao2023havatar} achieves fine-grained maintenance of tooth details, it struggles with hair details (the $1^{\text{th}}$ and $2^{\text{th}}$ row in Fig.~\ref{fig:self-compare}). The image quality of the results from PointAvatar~\cite{Zheng_2023_CVPR} is reduced due to the empty holes in dense point clouds (the $1^{\text{th}}$ row). The Gaussian-Head adopts the 3D Gaussian splats to address the artifacts from PointAvatar, but it still fails to represent the high-frequency details in torso texture and neck region (the $2^{\text{th}}$ and $3^{\text{th}}$ row). These results suggest that our method archives the best global image and local detail quality than the state-of-the-art methods, the local patch attention is still necessary for high-fidelity portrait avatars.

\vspace{-0.1cm}
\subsection{Qualitative Evaluation}
\label{Qualitative Evaluation}
\vspace{-0.1cm}

\noindent \textbf{Evaluation Settings.}  The qualitative evaluations aimed at elucidating the distinctions among baselines. We perform the self-/ and cross-/reenactment experiments on the videos from Dataset A and Dataset B.

\noindent \textbf{Evaluation Results.} Our qualitative evaluation highlights differences among various baseline methods in self-/ and cross-/ identity reconstructions (as shown in Fig.~\ref{fig:self-compare} and Fig.\ref{fig:cross-compare}). The GaussianHead~\cite{wang2024gaussianhead} exhibits facial details but unclear textures in torso areas due to geometric accuracy limitations. The PointAvatar~\cite{Zheng_2023_CVPR} heavily relies on the density of the points cloud and struggles with appearance recovery, especially around the tooth region. The HAvtar~\cite{zhao2023havatar} introduces the noise embedding of StyleGAN in the SR module, which will bring unexpected facial structure deformation. As the most advanced method, our method excels in self- / cross- reenactment quality and achieves the highest facial quality and motion accuracy.

\noindent \textbf{User Study.} We follow the procedure of Human Evaluation as in Deep Video Portrait~\cite{kim2018deep} and ER-NeRF~\cite{li2023efficient} to perform a user study of visual quality evaluation. We sample 60 videos (30 for self-reenactment and 30 for cross-reenactment) and invite 50 attendees on the Amazon Web Services (AWS) platform by Google questionnaire to assess the quality of sampled videos in several aspects. The Mean Opinion Scores (MOS) rating protocol is adopted for evaluation and the attendees are required to rate the sampled videos with the following questions:(1) MOS$_1$: \textit{``How about the image quality in the video?"}, (2) MOS$_2$: \textit{``How about the video realness?"}, (3) MOS$_3$:\textit{``Do you think the movement is synchronized between two heads?"} (this question is only for cross-reenactment) and (4) MOS$_4$: \textit{``How about the overall quality of the video?"}. The videos are shown in a random order, and each video is shown exactly once to assess the first impression of attendees. As shown in Table~\ref{table_1} (right), the HAvatar~\cite{zhao2023havatar} introduces adversarial training and improves the quality of image levels ($4.49 \text{ v.s. } 4.43$), but our method significantly better than it in video quality, motion accuracy and overall evaluation.

\begin{figure*}[t]
    \centering
    \includegraphics[width=1.\textwidth]{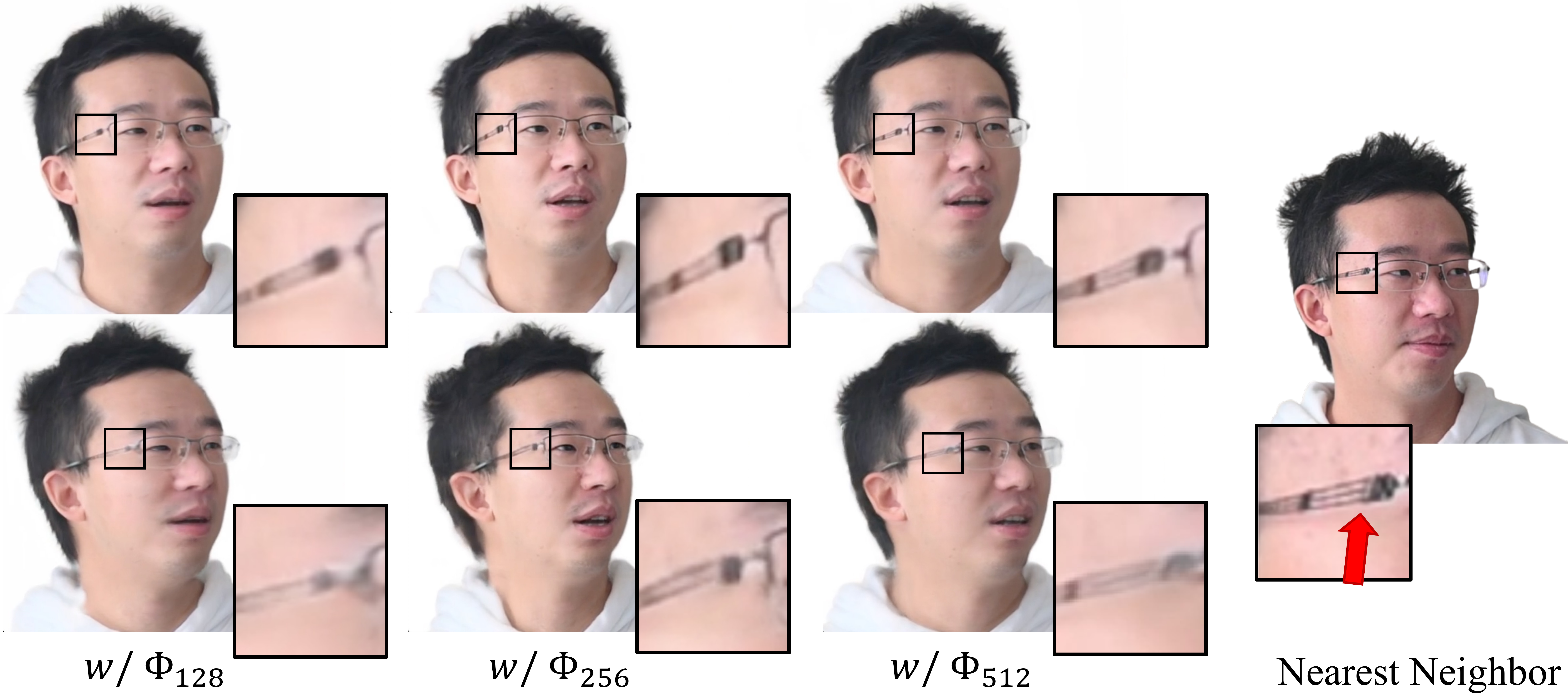}
    \vspace{-0.7cm}
    \caption{
        Visualization of different number of tri-planes. Two \textbf{novel-view reconstruction} results of same facial pose/expression are shown in each pair. The areas of interest have been zoomed in (with red arrow). We ($w/$ $\Phi_{512}$) exhibits best details, as evidenced by the glass textures ($w/$ $\Phi_{512}$ has three slices, others have two). The nearest neighbor shows the clearest ground truth of glass texture , which is from dataset.
    } \label{fig:tri-ab}
    \vspace{-.4cm}
\end{figure*}

\vspace{-0.3cm}
\subsection{Ablation Study}
\label{Ablation Study}
\vspace{-0.1cm}

\begin{figure}[t]
\centering
\begin{minipage}[b]{0.49\linewidth}
\centering
\includegraphics[width=6cm]{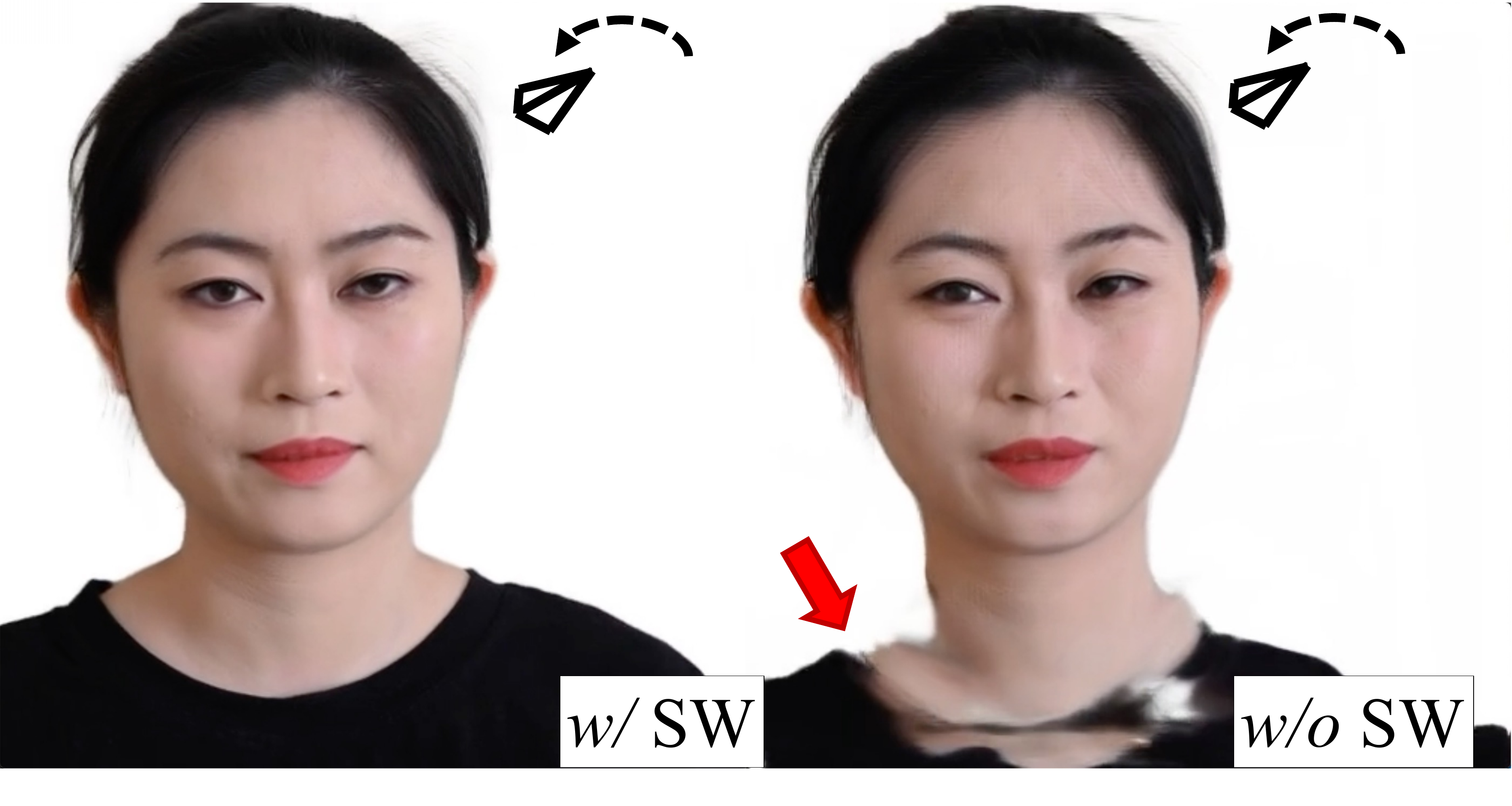}
\vspace{-.6cm}
\captionsetup{font={small,stretch=1}}
\caption{Visualization of improvement via geometry aware sliding window. We move camera to sample the canonical appearance from different viewpoints. The sliding window helps prevent artifacts from different views and movements.} \label{fig:ablationstudy}
\end{minipage}
\begin{minipage}[b]{0.49\linewidth}
    \centering
    \resizebox{1.0\textwidth}{!}{
    \begin{tabular}{cccc}
        \hlinew{1.15pt}
        \multirow{2}{*}{Methods} &F-LMD$\downarrow$  &PSNR$\uparrow$ &LPIPS$\downarrow$\\
        ~& &  & ($\times 10^{2}$)    \\
        \hline             
        $w/$ $\Phi_{128}$ & \textbf{2.35} & 25.50 & 8.80    \\
        $w/$ $\Phi_{256}$ & 2.55 & 26.17 & \underline{5.84}      \\
        $w/$ $\Phi_{128}$ + $w/o$ SW & \underline{2.47} & 24.70 & 10.5    \\
        $w/$ $\Phi_{256}$ + $w/o$ SW & 2.69 & 25.34 & 9.37   \\
        $w/$ $\Phi_{512}$ + $w/o$ SW & 2.69 & \underline{26.45} & 8.11 \\
        \rowcolor{mygray} Ours & 2.48 & \textbf{27.74} & \textbf{5.81}   \\
       \hlinew{1.15pt}
    \end{tabular}}
    \captionsetup{font={small,stretch=1}}
    \vspace{-0.2cm}
    \captionof{table}{Quantitative results of ablation study. The experiments are performed under self-reconstruction setting and Dataset A and B. The $w/o$ SW represents the ablation study without geometry-aware sliding window. The best are bold.}
    \label{tab:ablationstudy}
\end{minipage}
\vspace{-0.3cm}
\end{figure}

To validate the effectiveness of our method components, we focus on the improvement from Tri$^2$-plane and Geometry-Aware sliding window. We deactivate each of them (Fig.~\ref{fig:tri-ab}, Fig.~\ref{fig:ablationstudy}) and report results in Table~\ref{tab:ablationstudy}.

\noindent \textbf{Different number of tri-planes.} There are three different scale tri-plane in our Tri$^2$-plane. To validate the effectiveness of each tri-plane, we evaluate the Tri$^2$-plane with ablation configurations: i) replace the Tri$^2$-plane with only one tri-plane $\{\Phi_{128}\}$ (set to $w/$ $\Phi_{128}$ in Table~\ref{tab:ablationstudy}), ii) replace the Tri$^2$-plane with two tri-lane $\{\Phi_{128}+\Phi_{256}\}$ (set to $w/$ $\Phi_{256}$) and iii) the proposed Tri$^2$-plane (set as Ours). We finetune the SR module while maintaining an output scale of $512\times512$. As shown in Table~\ref{tab:ablationstudy}, the image quality is significantly improved, especially LPIPS with increasing number of triplanes. While the F-LMD is slightly reduced due to the fragmented patches by multiple triplanes. From $\Phi_{128}$ to $\Phi_{256}$, the image is divided from $1$ patch to $4$ patches and will lead the destruction of facial structure (multiple patches to form the whole image). The position of facial keypoints is modified, resulting in the decay of F-LMD ($2.35$ v.s. $2.55$). However, the introduction of 16 patches greatly improves the reconstruction precision and therefore makes an advancement than $\Phi_{256}$ ($2.55$ v.s. $2.48$). Additionally,  we provide visualization results in Fig.~\ref{fig:tri-ab}, highlighting the effectiveness of the cascaded tri-planes in capturing high-frequency features such as glass details (indicated by zoomed-in area). This reconstructed high frequency details are 3D consistent under different views, and will not switch with view directions. As shown in Fig.~\ref{fig:tri-ab}, even under different viewpoints, the three-piece glasses texture is well-preserved.

\noindent \textbf{The improvement via geometry aware sliding window.} We report the F-LMD, PSNR and LPIPS in the Table~\ref{tab:ablationstudy} to present results of with (set as Ours) and without (set as $w/o$ SW) geometry-aware sliding window.  As shown in Table~\ref{tab:ablationstudy}, the introduction of sliding window helps to improve the image quality on the different scale of rendered maps (\eg, $\Phi_{128}$: $25.50$ v.s. $24.70$, $\Phi_{256}$: $8.11$ v.s. $5.84$), which is a robustness improvement. However, it will take a minor decay of motion consistency (F-LMD: $2.48$ v.s. $2.55$), since the augment of translation ignores the head inherent rotation. The monocular videos usually lack of head rotation data, which biases the learnable network to translation instead of rotation. Visualization of improvement from sliding window are shown in Fig.~\ref{fig:ablationstudy}. We move the camera in circle to sample the canonical appearance from different viewpoints, it could be found that the quality of appearance $w/$ SW is better than $w/o$ SW, especially in the torso. The SW helps include more torso information during training, and those information are registered in the neural radiation field for the rendering with novel view or position.

\vspace{-0.5cm}
\section{Conclusion}
\label{Conclusion}
\vspace{-0.4cm}

In this work, we present a novel Feature Pyramid-based neural radiation field framework, Tri$^2$-plane, to reconstruct high-fidelity person-specific head avatars from a short monocular video. The proposed pyramid tri-plane feature spaces exhibit significantly improved representation capabilities, allowing for the faithful capture of high-frequency facial details. Moreover, we propose a simple yet effective geometry-aware sliding window to improve the 3D consistency. We should also acknowledge the limitations of our method. It relies on long-term videos for better performance, which are usually difficult to obtain in the wild.

\bibliographystyle{splncs04}
\bibliography{main}

\begin{thebibliography}{10}
\providecommand{\url}[1]{\texttt{#1}}
\providecommand{\urlprefix}{URL }
\providecommand{\doi}[1]{https://doi.org/#1}

\bibitem{abdal2019image2stylegan}
Abdal, R., Qin, Y., Wonka, P.: Image2stylegan: How to embed images into the stylegan latent space? In: Proceedings of the IEEE/CVF international conference on computer vision. pp. 4432--4441 (2019)

\bibitem{athar2022rignerf}
Athar, S., Xu, Z., Sunkavalli, K., Shechtman, E., Shu, Z.: Rignerf: Fully controllable neural 3d portraits. In: Proceedings of the IEEE/CVF conference on Computer Vision and Pattern Recognition. pp. 20364--20373 (2022)

\bibitem{bai2023high}
Bai, Y., Fan, Y., Wang, X., Zhang, Y., Sun, J., Yuan, C., Shan, Y.: High-fidelity facial avatar reconstruction from monocular video with generative priors. In: Proceedings of the IEEE/CVF Conference on Computer Vision and Pattern Recognition. pp. 4541--4551 (2023)

\bibitem{blanz2023morphable}
Blanz, V., Vetter, T.: A morphable model for the synthesis of 3d faces. In: Seminal Graphics Papers: Pushing the Boundaries, Volume 2, pp. 157--164 (2023)

\bibitem{chan2022efficient}
Chan, E.R., Lin, C.Z., Chan, M.A., Nagano, K., Pan, B., De~Mello, S., Gallo, O., Guibas, L.J., Tremblay, J., Khamis, S., et~al.: Efficient geometry-aware 3d generative adversarial networks. In: Proceedings of the IEEE/CVF Conference on Computer Vision and Pattern Recognition. pp. 16123--16133 (2022)

\bibitem{chan2021basicvsr}
Chan, K.C., Wang, X., Yu, K., Dong, C., Loy, C.C.: Basicvsr: The search for essential components in video super-resolution and beyond. In: Proceedings of the IEEE/CVF conference on computer vision and pattern recognition. pp. 4947--4956 (2021)

\bibitem{chan2022basicvsr++}
Chan, K.C., Zhou, S., Xu, X., Loy, C.C.: Basicvsr++: Improving video super-resolution with enhanced propagation and alignment. In: Proceedings of the IEEE/CVF conference on computer vision and pattern recognition. pp. 5972--5981 (2022)

\bibitem{chen2018lip}
Chen, L., Li, Z., Maddox, R.K., Duan, Z., Xu, C.: Lip movements generation at a glance. In: Proceedings of the European conference on computer vision (ECCV). pp. 520--535 (2018)

\bibitem{chung2018voxceleb2}
Chung, J.S., Nagrani, A., Zisserman, A.: Voxceleb2: Deep speaker recognition. arXiv preprint arXiv:1806.05622  (2018)

\bibitem{deng2019accurate}
Deng, Y., Yang, J., Xu, S., Chen, D., Jia, Y., Tong, X.: Accurate 3d face reconstruction with weakly-supervised learning: From single image to image set. In: Proceedings of the IEEE/CVF conference on computer vision and pattern recognition workshops. pp.~0--0 (2019)

\bibitem{dong2015image}
Dong, C., Loy, C.C., He, K., Tang, X.: Image super-resolution using deep convolutional networks. IEEE transactions on pattern analysis and machine intelligence  \textbf{38}(2),  295--307 (2015)

\bibitem{doukas2021head2head++}
Doukas, M.C., Koujan, M.R., Sharmanska, V., Roussos, A., Zafeiriou, S.: Head2head++: Deep facial attributes re-targeting. IEEE Transactions on Biometrics, Behavior, and Identity Science  \textbf{3}(1),  31--43 (2021)

\bibitem{gafni2021dynamic}
Gafni, G., Thies, J., Zollhofer, M., Nie{\ss}ner, M.: Dynamic neural radiance fields for monocular 4d facial avatar reconstruction. In: Proceedings of the IEEE/CVF Conference on Computer Vision and Pattern Recognition. pp. 8649--8658 (2021)

\bibitem{Gao2022nerfblendshape}
Gao, X., Zhong, C., Xiang, J., Hong, Y., Guo, Y., Zhang, J.: Reconstructing personalized semantic facial nerf models from monocular video. ACM Transactions on Graphics (Proceedings of SIGGRAPH Asia)  \textbf{41}(6) (2022). \doi{10.1145/3550454.3555501}

\bibitem{garrido2014automatic}
Garrido, P., Valgaerts, L., Rehmsen, O., Thormahlen, T., Perez, P., Theobalt, C.: Automatic face reenactment. In: Proceedings of the IEEE conference on computer vision and pattern recognition. pp. 4217--4224 (2014)

\bibitem{gerig2018morphable}
Gerig, T., Morel-Forster, A., Blumer, C., Egger, B., Luthi, M., Sch{\"o}nborn, S., Vetter, T.: Morphable face models-an open framework. In: 2018 13th IEEE International Conference on Automatic Face \& Gesture Recognition (FG 2018). pp. 75--82. IEEE (2018)

\bibitem{ghiasi2019fpn}
Ghiasi, G., Lin, T.Y., Le, Q.V.: Nas-fpn: Learning scalable feature pyramid architecture for object detection. In: Proceedings of the IEEE/CVF conference on computer vision and pattern recognition. pp. 7036--7045 (2019)

\bibitem{goodfellow2020generative}
Goodfellow, I., Pouget-Abadie, J., Mirza, M., Xu, B., Warde-Farley, D., Ozair, S., Courville, A., Bengio, Y.: Generative adversarial networks. Communications of the ACM  \textbf{63}(11),  139--144 (2020)

\bibitem{grassal2022neural}
Grassal, P.W., Prinzler, M., Leistner, T., Rother, C., Nie{\ss}ner, M., Thies, J.: Neural head avatars from monocular rgb videos. In: Proceedings of the IEEE/CVF Conference on Computer Vision and Pattern Recognition. pp. 18653--18664 (2022)

\bibitem{guo2018cnn}
Guo, Y., Cai, J., Jiang, B., Zheng, J., et~al.: Cnn-based real-time dense face reconstruction with inverse-rendered photo-realistic face images. IEEE transactions on pattern analysis and machine intelligence  \textbf{41}(6),  1294--1307 (2018)

\bibitem{guo2021ad}
Guo, Y., Chen, K., Liang, S., Liu, Y.J., Bao, H., Zhang, J.: Ad-nerf: Audio driven neural radiance fields for talking head synthesis. In: Proceedings of the IEEE/CVF International Conference on Computer Vision. pp. 5784--5794 (2021)

\bibitem{harkonen2020ganspace}
H{\"a}rk{\"o}nen, E., Hertzmann, A., Lehtinen, J., Paris, S.: Ganspace: Discovering interpretable gan controls. Advances in neural information processing systems  \textbf{33},  9841--9850 (2020)

\bibitem{he2017mask}
He, K., Gkioxari, G., Doll{\'a}r, P., Girshick, R.: Mask r-cnn. In: Proceedings of the IEEE international conference on computer vision. pp. 2961--2969 (2017)

\bibitem{hu2023tri}
Hu, W., Wang, Y., Ma, L., Yang, B., Gao, L., Liu, X., Ma, Y.: Tri-miprf: Tri-mip representation for efficient anti-aliasing neural radiance fields. In: Proceedings of the IEEE/CVF International Conference on Computer Vision. pp. 19774--19783 (2023)

\bibitem{hua2024finematch}
Hua, H., Shi, J., Kafle, K., Jenni, S., Zhang, D., Collomosse, J., Cohen, S., Luo, J.: Finematch: Aspect-based fine-grained image and text mismatch detection and correction. arXiv preprint arXiv:2404.14715  (2024)

\bibitem{hua2024v2xum}
Hua, H., Tang, Y., Xu, C., Luo, J.: V2xum-llm: Cross-modal video summarization with temporal prompt instruction tuning. arXiv preprint arXiv:2404.12353  (2024)

\bibitem{huang2018introduction}
Huang, H., Yu, P.S., Wang, C.: An introduction to image synthesis with generative adversarial nets. arXiv preprint arXiv:1803.04469  (2018)

\bibitem{huang2023boosting}
Huang, Y., Dong, H., Pan, J., Zhu, C., Liang, B., Guo, Y., Liu, D., Fu, L., Wang, F.: Boosting video super resolution with patch-based temporal redundancy optimization. In: International Conference on Artificial Neural Networks. pp. 362--375. Springer (2023)

\bibitem{johnson2016perceptual}
Johnson, J., Alahi, A., Fei-Fei, L.: Perceptual losses for real-time style transfer and super-resolution. In: Computer Vision--ECCV 2016: 14th European Conference, Amsterdam, The Netherlands, October 11-14, 2016, Proceedings, Part II 14. pp. 694--711. Springer (2016)

\bibitem{karras2019style}
Karras, T., Laine, S., Aila, T.: A style-based generator architecture for generative adversarial networks. In: Proceedings of the IEEE/CVF conference on computer vision and pattern recognition. pp. 4401--4410 (2019)

\bibitem{karras2020analyzing}
Karras, T., Laine, S., Aittala, M., Hellsten, J., Lehtinen, J., Aila, T.: Analyzing and improving the image quality of stylegan. In: Proceedings of the IEEE/CVF conference on computer vision and pattern recognition. pp. 8110--8119 (2020)

\bibitem{kerbl20233d}
Kerbl, B., Kopanas, G., Leimk{\"u}hler, T., Drettakis, G.: 3d gaussian splatting for real-time radiance field rendering. ACM Transactions on Graphics  \textbf{42}(4) (2023)

\bibitem{kim2018deep}
Kim, H., Garrido, P., Tewari, A., Xu, W., Thies, J., Niessner, M., P{\'e}rez, P., Richardt, C., Zollh{\"o}fer, M., Theobalt, C.: Deep video portraits. ACM transactions on graphics (TOG)  \textbf{37}(4),  1--14 (2018)

\bibitem{kingma2014adam}
Kingma, D.P., Ba, J.: Adam: A method for stochastic optimization. arXiv preprint arXiv:1412.6980  (2014)

\bibitem{kirschstein2023nersemble}
Kirschstein, T., Qian, S., Giebenhain, S., Walter, T., Nie\ss{}ner, M.: Nersemble: Multi-view radiance field reconstruction of human heads. ACM Trans. Graph.  \textbf{42}(4) (jul 2023). \doi{10.1145/3592455}, \url{https://doi.org/10.1145/3592455}

\bibitem{kocabas2023hugs}
Kocabas, M., Chang, J.H.R., Gabriel, J., Tuzel, O., Ranjan, A.: Hugs: Human gaussian splats. arXiv preprint arXiv:2311.17910  (2023)

\bibitem{kong2018deep}
Kong, T., Sun, F., Tan, C., Liu, H., Huang, W.: Deep feature pyramid reconfiguration for object detection. In: Proceedings of the European conference on computer vision (ECCV). pp. 169--185 (2018)

\bibitem{koujan2020head2head}
Koujan, M.R., Doukas, M.C., Roussos, A., Zafeiriou, S.: Head2head: Video-based neural head synthesis. In: 2020 15th IEEE International Conference on Automatic Face and Gesture Recognition (FG 2020). pp. 16--23. IEEE (2020)

\bibitem{li2023efficient}
Li, J., Zhang, J., Bai, X., Zhou, J., Gu, L.: Efficient region-aware neural radiance fields for high-fidelity talking portrait synthesis. In: Proceedings of the IEEE/CVF International Conference on Computer Vision. pp. 7568--7578 (2023)

\bibitem{li2024gaussianbody}
Li, M., Yao, S., Xie, Z., Chen, K., Jiang, Y.G.: Gaussianbody: Clothed human reconstruction via 3d gaussian splatting. arXiv preprint arXiv:2401.09720  (2024)

\bibitem{FLAME:SiggraphAsia2017}
Li, T., Bolkart, T., Black, M.J., Li, H., Romero, J.: Learning a model of facial shape and expression from {4D} scans. ACM Transactions on Graphics, (Proc. SIGGRAPH Asia)  \textbf{36}(6),  194:1--194:17 (2017), \url{https://doi.org/10.1145/3130800.3130813}

\bibitem{li2023one}
Li, W., Zhang, L., Wang, D., Zhao, B., Wang, Z., Chen, M., Zhang, B., Wang, Z., Bo, L., Li, X.: One-shot high-fidelity talking-head synthesis with deformable neural radiance field. In: Proceedings of the IEEE/CVF Conference on Computer Vision and Pattern Recognition. pp. 17969--17978 (2023)

\bibitem{li2023animatable}
Li, Z., Zheng, Z., Wang, L., Liu, Y.: Animatable gaussians: Learning pose-dependent gaussian maps for high-fidelity human avatar modeling. arXiv preprint arXiv:2311.16096  (2023)

\bibitem{lin2017feature}
Lin, T.Y., Doll{\'a}r, P., Girshick, R., He, K., Hariharan, B., Belongie, S.: Feature pyramid networks for object detection. In: Proceedings of the IEEE conference on computer vision and pattern recognition. pp. 2117--2125 (2017)

\bibitem{lin2017focal}
Lin, T.Y., Goyal, P., Girshick, R., He, K., Doll{\'a}r, P.: Focal loss for dense object detection. In: Proceedings of the IEEE international conference on computer vision. pp. 2980--2988 (2017)

\bibitem{liu2018path}
Liu, S., Qi, L., Qin, H., Shi, J., Jia, J.: Path aggregation network for instance segmentation. In: Proceedings of the IEEE conference on computer vision and pattern recognition. pp. 8759--8768 (2018)

\bibitem{lugaresi2019mediapipe}
Lugaresi, C., Tang, J., Nash, H., McClanahan, C., Uboweja, E., Hays, M., Zhang, F., Chang, C.L., Yong, M.G., Lee, J., et~al.: Mediapipe: A framework for building perception pipelines. arXiv preprint arXiv:1906.08172  (2019)

\bibitem{ma2023otavatar}
Ma, Z., Zhu, X., Qi, G.J., Lei, Z., Zhang, L.: Otavatar: One-shot talking face avatar with controllable tri-plane rendering. In: Proceedings of the IEEE/CVF Conference on Computer Vision and Pattern Recognition. pp. 16901--16910 (2023)

\bibitem{mathieu2015deep}
Mathieu, M., Couprie, C., LeCun, Y.: Deep multi-scale video prediction beyond mean square error. arXiv preprint arXiv:1511.05440  (2015)

\bibitem{mildenhall2021nerf}
Mildenhall, B., Srinivasan, P.P., Tancik, M., Barron, J.T., Ramamoorthi, R., Ng, R.: Nerf: Representing scenes as neural radiance fields for view synthesis. Communications of the ACM  \textbf{65}(1),  99--106 (2021)

\bibitem{mueller2022instant}
M\"uller, T., Evans, A., Schied, C., Keller, A.: Instant neural graphics primitives with a multiresolution hash encoding. ACM Trans. Graph.  \textbf{41}(4),  102:1--102:15 (Jul 2022). \doi{10.1145/3528223.3530127}, \url{https://doi.org/10.1145/3528223.3530127}

\bibitem{nagrani2020voxceleb}
Nagrani, A., Chung, J.S., Xie, W., Zisserman, A.: Voxceleb: Large-scale speaker verification in the wild. Computer Speech \& Language  \textbf{60},  101027 (2020)

\bibitem{pang2019libra}
Pang, J., Chen, K., Shi, J., Feng, H., Ouyang, W., Lin, D.: Libra r-cnn: Towards balanced learning for object detection. In: Proceedings of the IEEE/CVF conference on computer vision and pattern recognition. pp. 821--830 (2019)

\bibitem{park2021nerfies}
Park, K., Sinha, U., Barron, J.T., Bouaziz, S., Goldman, D.B., Seitz, S.M., Martin-Brualla, R.: Nerfies: Deformable neural radiance fields. In: Proceedings of the IEEE/CVF International Conference on Computer Vision. pp. 5865--5874 (2021)

\bibitem{qian2023gaussianavatars}
Qian, S., Kirschstein, T., Schoneveld, L., Davoli, D., Giebenhain, S., Nie{\ss}ner, M.: Gaussianavatars: Photorealistic head avatars with rigged 3d gaussians. arXiv preprint arXiv:2312.02069  (2023)

\bibitem{raj2021pva}
Raj, A., Zollhoefer, M., Simon, T., Saragih, J., Saito, S., Hays, J., Lombardi, S.: Pva: Pixel-aligned volumetric avatars. arXiv preprint arXiv:2101.02697  (2021)

\bibitem{richardson2021encoding}
Richardson, E., Alaluf, Y., Patashnik, O., Nitzan, Y., Azar, Y., Shapiro, S., Cohen-Or, D.: Encoding in style: a stylegan encoder for image-to-image translation. In: Proceedings of the IEEE/CVF conference on computer vision and pattern recognition. pp. 2287--2296 (2021)

\bibitem{roich2022pivotal}
Roich, D., Mokady, R., Bermano, A.H., Cohen-Or, D.: Pivotal tuning for latent-based editing of real images. ACM Transactions on graphics (TOG)  \textbf{42}(1),  1--13 (2022)

\bibitem{ronneberger2015u}
Ronneberger, O., Fischer, P., Brox, T.: U-net: Convolutional networks for biomedical image segmentation. In: Medical Image Computing and Computer-Assisted Intervention--MICCAI 2015: 18th International Conference, Munich, Germany, October 5-9, 2015, Proceedings, Part III 18. pp. 234--241. Springer (2015)

\bibitem{saito2024rgca}
Saito, S., Schwartz, G., Simon, T., Li, J., Nam, G.: Relightable gaussian codec avatars. In: CVPR (2024)

\bibitem{shen2021closed}
Shen, Y., Zhou, B.: Closed-form factorization of latent semantics in gans. In: Proceedings of the IEEE/CVF conference on computer vision and pattern recognition. pp. 1532--1540 (2021)

\bibitem{shi20223d}
Shi, Z., Shen, Y., Zhu, J., Yeung, D.Y., Chen, Q.: 3d-aware indoor scene synthesis with depth priors. In: European Conference on Computer Vision. pp. 406--422. Springer (2022)

\bibitem{song2022adaptive}
Song, L., Fang, Z., Li, X., Dong, X., Jin, Z., Chen, Y., Lyu, S.: Adaptive face forgery detection in cross domain. In: European Conference on Computer Vision. pp. 467--484. Springer (2022)

\bibitem{song2022face}
Song, L., Li, X., Fang, Z., Jin, Z., Chen, Y., Xu, C.: Face forgery detection via symmetric transformer. In: Proceedings of the 30th ACM international conference on multimedia. pp. 4102--4111 (2022)

\bibitem{song2021tacr}
Song, L., Liu, B., Yin, G., Dong, X., Zhang, Y., Bai, J.X.: Tacr-net: editing on deep video and voice portraits. In: Proceedings of the 29th ACM International Conference on Multimedia. pp. 478--486 (2021)

\bibitem{song2021talking}
Song, L., Liu, B., Yu, N.: Talking face video generation with editable expression. In: Image and Graphics: 11th International Conference, ICIG 2021, Haikou, China, August 6--8, 2021, Proceedings, Part III 11. pp. 753--764. Springer (2021)

\bibitem{song2024adaptive}
Song, L., Liu, P., Yin, G., Xu, C.: Adaptive super resolution for one-shot talking-head generation. In: ICASSP 2024-2024 IEEE International Conference on Acoustics, Speech and Signal Processing (ICASSP). pp. 4115--4119. IEEE (2024)

\bibitem{song2023emotional}
Song, L., Yin, G., Jin, Z., Dong, X., Xu, C.: Emotional listener portrait: Neural listener head generation with emotion. In: Proceedings of the IEEE/CVF International Conference on Computer Vision. pp. 20839--20849 (2023)

\bibitem{song2021fsft}
Song, L., Yin, G., Liu, B., Zhang, Y., Yu, N.: Fsft-net: face transfer video generation with few-shot views. In: 2021 IEEE International Conference on Image Processing (ICIP). pp. 3582--3586. IEEE (2021)

\bibitem{su2021nerf}
Su, S.Y., Yu, F., Zollh{\"o}fer, M., Rhodin, H.: A-nerf: Articulated neural radiance fields for learning human shape, appearance, and pose. Advances in Neural Information Processing Systems  \textbf{34},  12278--12291 (2021)

\bibitem{sun2022ide}
Sun, J., Wang, X., Shi, Y., Wang, L., Wang, J., Liu, Y.: Ide-3d: Interactive disentangled editing for high-resolution 3d-aware portrait synthesis. ACM Transactions on Graphics (ToG)  \textbf{41}(6),  1--10 (2022)

\bibitem{sun2023next3d}
Sun, J., Wang, X., Wang, L., Li, X., Zhang, Y., Zhang, H., Liu, Y.: Next3d: Generative neural texture rasterization for 3d-aware head avatars. In: Proceedings of the IEEE/CVF Conference on Computer Vision and Pattern Recognition. pp. 20991--21002 (2023)

\bibitem{sun2022fenerf}
Sun, J., Wang, X., Zhang, Y., Li, X., Zhang, Q., Liu, Y., Wang, J.: Fenerf: Face editing in neural radiance fields. In: Proceedings of the IEEE/CVF Conference on Computer Vision and Pattern Recognition. pp. 7672--7682 (2022)

\bibitem{tan2020efficientdet}
Tan, M., Pang, R., Le, Q.V.: Efficientdet: Scalable and efficient object detection. In: Proceedings of the IEEE/CVF conference on computer vision and pattern recognition. pp. 10781--10790 (2020)

\bibitem{tang2022real}
Tang, J., Wang, K., Zhou, H., Chen, X., He, D., Hu, T., Liu, J., Zeng, G., Wang, J.: Real-time neural radiance talking portrait synthesis via audio-spatial decomposition. arXiv preprint arXiv:2211.12368  (2022)

\bibitem{tang2023video}
Tang, Y., Bi, J., Xu, S., Song, L., Liang, S., Wang, T., Zhang, D., An, J., Lin, J., Zhu, R., et~al.: Video understanding with large language models: A survey. arXiv preprint arXiv:2312.17432  (2023)

\bibitem{tang2024avicuna}
Tang, Y., Shimada, D., Bi, J., Xu, C.: Avicuna: Audio-visual llm with interleaver and context-boundary alignment for temporal referential dialogue. arXiv preprint arXiv:2403.16276  (2024)

\bibitem{Tang_2022_ACCV}
Tang, Y., Xu, S., Wang, T., Lin, Q., Lu, Q., Zheng, F.: Multi-modal segment assemblage network for ad video editing with importance-coherence reward. In: Proceedings of the Asian Conference on Computer Vision (ACCV). pp. 3519--3535 (December 2022)

\bibitem{teotia2023hq3davatar}
Teotia, K., Pan, X., Kim, H., Garrido, P., Elgharib, M., Theobalt, C., et~al.: Hq3davatar: High quality controllable 3d head avatar. arXiv preprint arXiv:2303.14471  (2023)

\bibitem{thies2020neural}
Thies, J., Elgharib, M., Tewari, A., Theobalt, C., Nie{\ss}ner, M.: Neural voice puppetry: Audio-driven facial reenactment. In: Computer Vision--ECCV 2020: 16th European Conference, Glasgow, UK, August 23--28, 2020, Proceedings, Part XVI 16. pp. 716--731. Springer (2020)

\bibitem{thies2016facevr}
Thies, J., Zollh{\"o}fer, M., Stamminger, M., Theobalt, C., Nie{\ss}ner, M.: Facevr: Real-time facial reenactment and eye gaze control in virtual reality. arXiv preprint arXiv:1610.03151  (2016)

\bibitem{trevithick2023}
Trevithick, A., Chan, M., Stengel, M., Chan, E.R., Liu, C., Yu, Z., Khamis, S., Chandraker, M., Ramamoorthi, R., Nagano, K.: Real-time radiance fields for single-image portrait view synthesis. In: ACM Transactions on Graphics (SIGGRAPH) (2023)

\bibitem{voynov2020unsupervised}
Voynov, A., Babenko, A.: Unsupervised discovery of interpretable directions in the gan latent space. In: International conference on machine learning. pp. 9786--9796. PMLR (2020)

\bibitem{wang2024gaussianhead}
Wang, J., Xie, J.C., Li, X., Xu, F., Pun, C.M., Gao, H.: Gaussianhead: High-fidelity head avatars with learnable gaussian derivation (2024)

\bibitem{wang2022faceverse}
Wang, L., Chen, Z., Yu, T., Ma, C., Li, L., Liu, Y.: Faceverse: a fine-grained and detail-controllable 3d face morphable model from a hybrid dataset. In: Proceedings of the IEEE/CVF conference on computer vision and pattern recognition. pp. 20333--20342 (2022)

\bibitem{wang2023styleavatar}
Wang, L., Zhao, X., Sun, J., Zhang, Y., Zhang, H., Yu, T., Liu, Y.: Styleavatar: Real-time photo-realistic portrait avatar from a single video. arXiv preprint arXiv:2305.00942  (2023)

\bibitem{xu2023gaussian}
Xu, Y., Chen, B., Li, Z., Zhang, H., Wang, L., Zheng, Z., Liu, Y.: Gaussian head avatar: Ultra high-fidelity head avatar via dynamic gaussians. arXiv preprint arXiv:2312.03029  (2023)

\bibitem{xu2023latentavatar}
Xu, Y., Zhang, H., Wang, L., Zhao, X., Huang, H., Qi, G., Liu, Y.: Latentavatar: Learning latent expression code for expressive neural head avatar. arXiv preprint arXiv:2305.01190  (2023)

\bibitem{yang2022Vtoonify}
Yang, S., Jiang, L., Liu, Z., Loy, C.C.: Vtoonify: Controllable high-resolution portrait video style transfer. ACM Transactions on Graphics (TOG)  \textbf{41}(6),  1--15 (2022). \doi{10.1145/3550454.3555437}

\bibitem{yu2023nofa}
Yu, W., Fan, Y., Zhang, Y., Wang, X., Yin, F., Bai, Y., Cao, Y.P., Shan, Y., Wu, Y., Sun, Z., et~al.: Nofa: Nerf-based one-shot facial avatar reconstruction. In: ACM SIGGRAPH 2023 Conference Proceedings. pp. 1--12 (2023)

\bibitem{yu2024promptfix}
Yu, Y., Zeng, Z., Hua, H., Fu, J., Luo, J.: Promptfix: You prompt and we fix the photo. arXiv preprint arXiv:2405.16785  (2024)

\bibitem{zhang2020feature}
Zhang, D., Zhang, H., Tang, J., Wang, M., Hua, X., Sun, Q.: Feature pyramid transformer. In: Computer Vision--ECCV 2020: 16th European Conference, Glasgow, UK, August 23--28, 2020, Proceedings, Part XXVIII 16. pp. 323--339. Springer (2020)

\bibitem{zhang2018unreasonable}
Zhang, R., Isola, P., Efros, A.A., Shechtman, E., Wang, O.: The unreasonable effectiveness of deep features as a perceptual metric. In: Proceedings of the IEEE conference on computer vision and pattern recognition. pp. 586--595 (2018)

\bibitem{zhao2021graphfpn}
Zhao, G., Ge, W., Yu, Y.: Graphfpn: Graph feature pyramid network for object detection. In: Proceedings of the IEEE/CVF international conference on computer vision. pp. 2763--2772 (2021)

\bibitem{zhao2023havatar}
Zhao, X., Wang, L., Sun, J., Zhang, H., Suo, J., Liu, Y.: Havatar: High-fidelity head avatar via facial model conditioned neural radiance field. ACM Transactions on Graphics  (2023)

\bibitem{zheng2022avatar}
Zheng, Y., Abrevaya, V.F., B{\"u}hler, M.C., Chen, X., Black, M.J., Hilliges, O.: Im avatar: Implicit morphable head avatars from videos. In: Proceedings of the IEEE/CVF Conference on Computer Vision and Pattern Recognition. pp. 13545--13555 (2022)

\bibitem{Zheng_2023_CVPR}
Zheng, Y., Yifan, W., Wetzstein, G., Black, M.J., Hilliges, O.: Pointavatar: Deformable point-based head avatars from videos. In: Proceedings of the IEEE/CVF Conference on Computer Vision and Pattern Recognition (CVPR). pp. 21057--21067 (June 2023)

\bibitem{Zielonka2023Drivable3D}
Zielonka, W., Bagautdinov, T., Saito, S., Zollhöfer, M., Thies, J., Romero, J.: Drivable 3d gaussian avatars  (2023)

\bibitem{zielonka2023instant}
Zielonka, W., Bolkart, T., Thies, J.: Instant volumetric head avatars. In: Proceedings of the IEEE/CVF Conference on Computer Vision and Pattern Recognition. pp. 4574--4584 (2023)

\end{thebibliography}

\clearpage

\section{Appendix}
\label{sec:supple}

\subsection{Related Works}
\label{Related Works}

\noindent \textbf{Deformable Neural Radiance Fields.} The deformable-NeRF~\cite{athar2022rignerf,gafni2021dynamic,park2021nerfies,su2021nerf} is for dynamic scenes reenactment by providing as input a time component, some of imposing temporal constraints or scene flow, while the others~\cite{thies2020neural,guo2021ad,raj2021pva,sun2023next3d} using a canonical frame/template to simulate motion by moving the camera position. Some notable work, such as INSTA~\cite{zielonka2023instant} adopts a bounding volume hierarchy to accelerate point searching along rays. RAD-NeRF~\cite{tang2022real} and ER-NeRF~\cite{li2023efficient}, following Instant-NGP~\cite{mueller2022instant}, utilized a set of hash tables to reduce the number of feature grids.

\noindent \textbf{Facial Video Animation}. The previous animation works employ morphable model reconstruction, forward rendering with optimized textures~\cite{garrido2014automatic, thies2016facevr}, and image or video synthesis~\cite{hua2024finematch,Tang_2022_ACCV,hua2024v2xum} through deep neural network~\cite{karras2019style, huang2018introduction,tang2024avicuna,tang2023video,yu2024promptfix}. Some of them~\cite{kim2018deep, koujan2020head2head, doukas2021head2head++,wang2023styleavatar} utilizes rendered correspondence maps in conjunction with image-to-image translation networks to produce highly realistic videos. Other methods~\cite{grassal2022neural, zheng2022avatar, Zheng_2023_CVPR} use implicit geometry and texture models to overcome the limitations of texture quality by image translation network. 

\noindent \textbf{3D Gaussian-Splat Rendering}. Recently, 3D Gaussian splatting~\cite{kerbl20233d} shows its superior performance in synthesis quality and rendering speed. It leverage the point elements as a discrete and unstructured representation to fit geometry with arbitrary topology~\cite{xu2023gaussian}. Some approaches~\cite{xu2023latentavatar,qian2023gaussianavatars,Zielonka2023Drivable3D,li2024gaussianbody,li2023animatable,kocabas2023hugs} extend Gaussian representation to multi-view avatar reconstruction. However, these methods can not be migrated to the monocular head avatar reconstruction setting.

\subsection{Geometry-Aware Sliding Window}

\begin{figure*}[t]
    \centering
    \includegraphics[width=1.\textwidth]{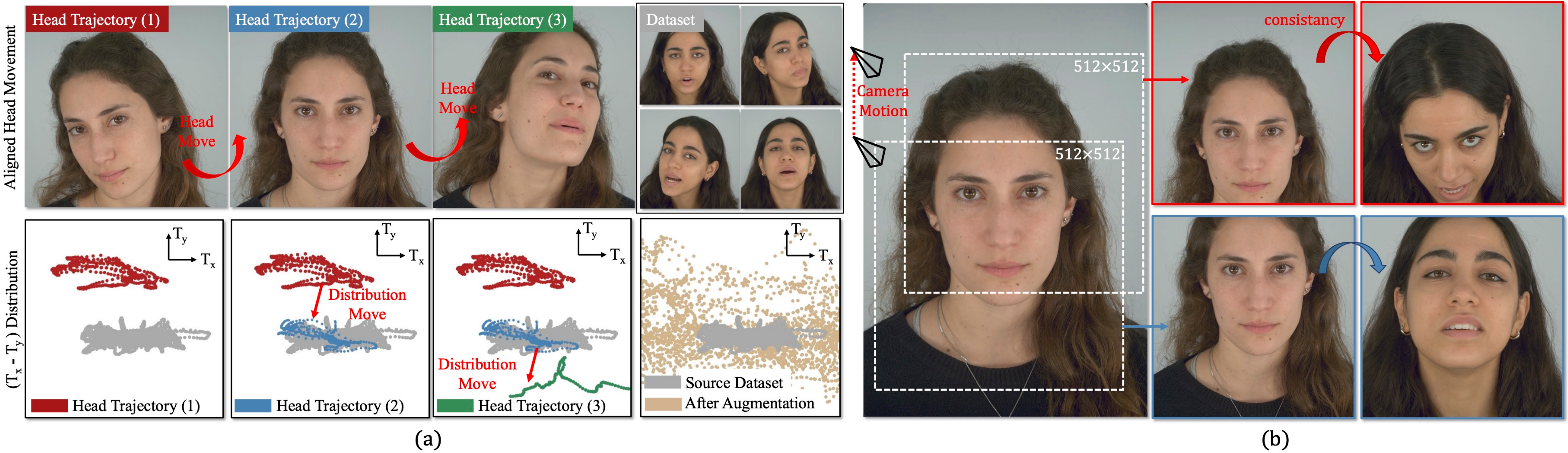}
    \vspace{-0.7cm}
    \caption{
        (1) Left: The upper part demonstrates head movement trajectories alongside head position samples from the aligned training set. The lower part presents the distribution of translation values ($\mathbf{T}_x$, $\mathbf{T}_y$), with trajectory distributions shown in red, blue, and green, contrasted against the training set's gray points. The augmented distribution, encompassing these trajectories, is denoted with yellow points. (2) The sliding window simulation of head movement. It expands the torso in larger range. 
    } \label{fig:appendix-2}
    \vspace{-.4cm}
\end{figure*}

The illustrations of geometry-aware sliding window in left and right part of Fig.~\ref{fig:appendix-2}. In the left, the sliding windows extend the distribution range withiin the aligned facial dataset. The augmentation addresses the distribution distortions arising from head trajectory movements. In the right part in Fig.~\ref{fig:appendix-2}, we demonstrate simulating realistic head diversity motion via sliding window. The upward head movement matches the downward sliding window translation, and vice versa.

\vspace{-0.2cm}
\subsection{Experiments}
\vspace{-0.1cm}

\begin{figure*}[t]
    \centering
    \includegraphics[width=1.\linewidth]{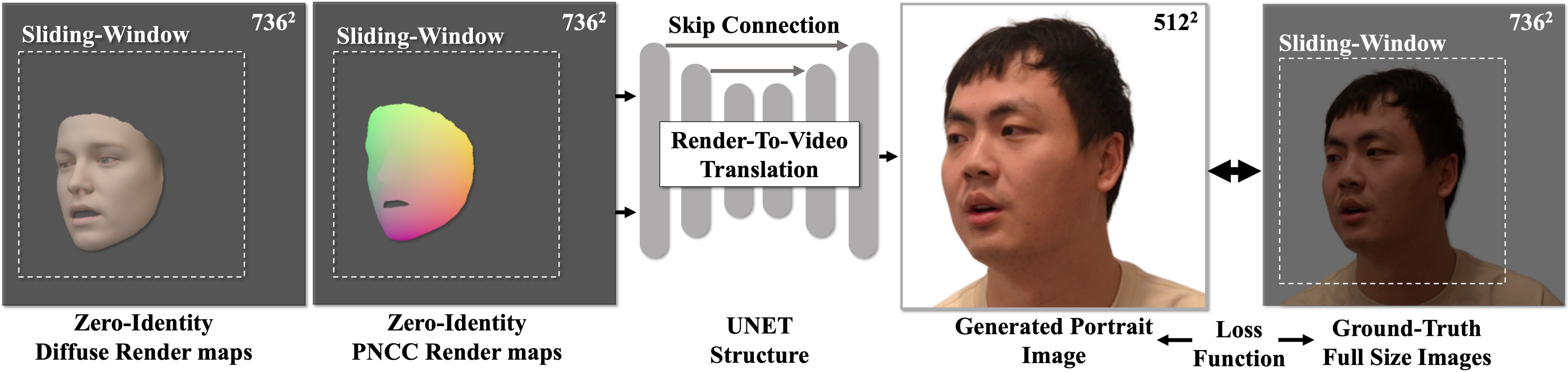}
    \caption{
    \textbf{Overview of DVP in our baselines.} We employ the above structure for the DVP~\cite{kim2018deep} in experiments. The Render-To-Video Translation network takes the inputs (\textbf{Zero-Identity Diffuse Render maps} and \textbf{Zero-Identity PNCC Render maps}) from $6\times512\times512$ to the outputs (\textbf{Generated Portrait Images}) with $3\times512\times512$. Additionally, we adopt the directly sliding windows to improve performance. Each size of input and output is shown in above. The losses are calculated in the corresponding area of the sliding window ($3\times512\times512$) in the ground-truth images (\textbf{Ground-Truth Full Size Images}, $3\times736\times736$) and outputs to optimize the whole network.
    } \label{fig:dvp-pipeline}
\vspace{-0.2cm}
\end{figure*}

\noindent \textbf{Extend Baseline.} We extend the evaluation on six more state-of-the-art methods, including:
\begin{itemize}
\item IMAvatar~\cite{zheng2022avatar}: We train the model for each video with 60 epochs.  The image size in it is rescaled to $512\times512$, while the original size is $256\times256$.
\item Instant-Avatar (INSTA)~\cite{zielonka2023instant}: We take training code with PyTorch.  The Torso part is retained for fair comparison. 
\item RAD-NeRF~\cite{tang2022real} and ER-NeRF~\cite{li2023efficient}: We follow the official pipeline provided by ER-NeRF~\cite{li2023efficient} and RAD-NeRF~\cite{tang2022real}. However, their initial setting is audio-driven facial animation, which is different from ours. We replace the audio features with the expression coefficients in the inputs. The model is trained on each video with 60 epochs on face and 40 epochs on torso.  
\item Deep Video Portrait (DVP)~\cite{kim2018deep}: A foundational model for current methods~\cite{doukas2021head2head++,wang2023styleavatar}, reimplemented at a resolution of $512\times512$ with sliding windows in our experiments. Since there is no source code provided by the DVP~\cite{kim2018deep}, we configure the data according to the face-tracking method in our work. We find that the sliding window will significantly improve the robustness, and apply it in reimplementation. It is not a geometry-aware sliding window, it is directly aligned on image level. The pipeline of the DVP in baseline methods is shown in Fig.~\ref{fig:dvp-pipeline}. We train the model with batch size $16$. 
\item Style-Avatar~\cite{wang2023styleavatar}: We strictly follow the official process provided by StyleAvatar~\cite{wang2023styleavatar} and using FaceVerse~\cite{wang2022faceverse} to extract texture and uv maps. 
\end{itemize}
These baselines serve as a supplement to the main paper. Together with the three baselines in the main paper, which cover NeRF, tri-plane, Generative Adversarial Networks~\cite{goodfellow2020generative} (GAN), and Gaussian-splatting backbones. However, there are still many significant methods that are not included, such as \cite{ma2023otavatar, sun2023next3d, bai2023high, xu2023latentavatar, song2021talking}, which due to gaps in research topics~\cite{ma2023otavatar, sun2023next3d, bai2023high, xu2023latentavatar,yu2023nofa} and without open-source code.

\noindent \textbf{Metrics Detail.} We further define our metrics in detail, as a supplement to the main paper. Before we start, we declare the parameters that need to be used, the generated portrait images $\textbf{G}'_{j}$ and the ground-truth portrait images $\textbf{G}_{j}$.\par

\noindent \textit{Full Face Landmark (F-LMD)}. The F-LMD is the coordinate facial landmark distance between the full artistic face and source face via MediaPipe~\cite{lugaresi2019mediapipe} keypoints detector. It noted as $\text{LMD}(\cdot)$ for each portrait image landmarks, as: 
\begin{equation}
\text{F-LMD} = \sum ||\text{LMD}(\textbf{G}_{j}) - \text{LMD}(\textbf{G}'_{j})||_1,
\label{AKD}
\end{equation}
the number of landmarks applied in the F-LMD is $68$, and we use the $l_1$ distance between the detected keypoints from the source and generated images. \par

\noindent \textit{The Sharpness Difference (SD)}. The SD~\cite{mathieu2015deep} is defined as following: 
\begin{equation}
\footnotesize
\text{SD} = 10\text{log}_{10} \frac{N \cdot \text{max}^2_{\textbf{G}'_{j}}}{  \sum_{m,n} | (\nabla_{m} \textbf{G}'_{j} + \nabla_{n} \textbf{G}'_{j}) -  (\nabla_{m} \textbf{G}_{j} + \nabla_{n} \textbf{G}_{j}) | }   ,
\label{FD}
\end{equation}
where $\nabla_{m} \textbf{G}'_{j} = |\textbf{G}'_{j}[m,n] - \textbf{G}'_{j}[-1,n]|$ and $\nabla_{n} \textbf{G}'_{j} = |\textbf{G}'_{j}[m,n] - \textbf{G}'_{j}[m,n-1]|$. The $\textbf{G}'_{j}[m,n]$ represents the value of the pixel at each position. The $max_{\textbf{G}'_{j}}$ is the maximum possible pixel value of the image.

\noindent \textit{Peak Signal-to-Noise Ratio (PSNR)}. The PSNR is a standard metric used to measure the quality of reconstruction of lossy transformation processes, particularly in the field of image processing. It is defined as follows:
\begin{equation}
\text{PSNR} = 20 \cdot \log_{10}\left(\frac{\text{max}_{\textbf{G}'_{j}}}{||\textbf{G}'_{j} - \textbf{G}_{j}||_2 }\right),
\label{PSNR}
\end{equation}
where $\text{max}_{\textbf{G}'_{j}}$ is the maximum possible pixel value of the image, and $||\cdot||_2$ denotes the Mean Squared Error between the ground-truth $\textbf{G}_{j}$ and the generated image $\textbf{G}'_{j}$. The PSNR is typically expressed in terms of the logarithmic decibel scale.

\noindent \textit{Learned Perceptual Image Patch Similarity (LPIPS)}. The LPIPS metric quantifies perceptual similarity between two images using deep learning features. It is expressed as:
\begin{equation}
\footnotesize
\text{LPIPS} = \sum_{H_lW_l} \frac{1}{H_l W_l} \sum_{h,w}1 - \text{cos}(\phi_l(\textbf{G}_{j}[h,w]), \phi_l(\textbf{G}'_{j}[h,w])),
\label{LPIPS}
\end{equation}
where $\phi_l(\cdot)$ represents the feature map extracted by a pre-trained deep neural network at layer $l$, and $(h, w)$ are spatial locations in these feature maps. $H_l$ and $W_l$ are the height and width of the feature map at layer $l$, respectively. The LPIPS metric provides a more perceptually-aligned measure of similarity than traditional pixel-based metrics.

\begin{figure*}[t]
    \centering
    \includegraphics[width=1.\linewidth]{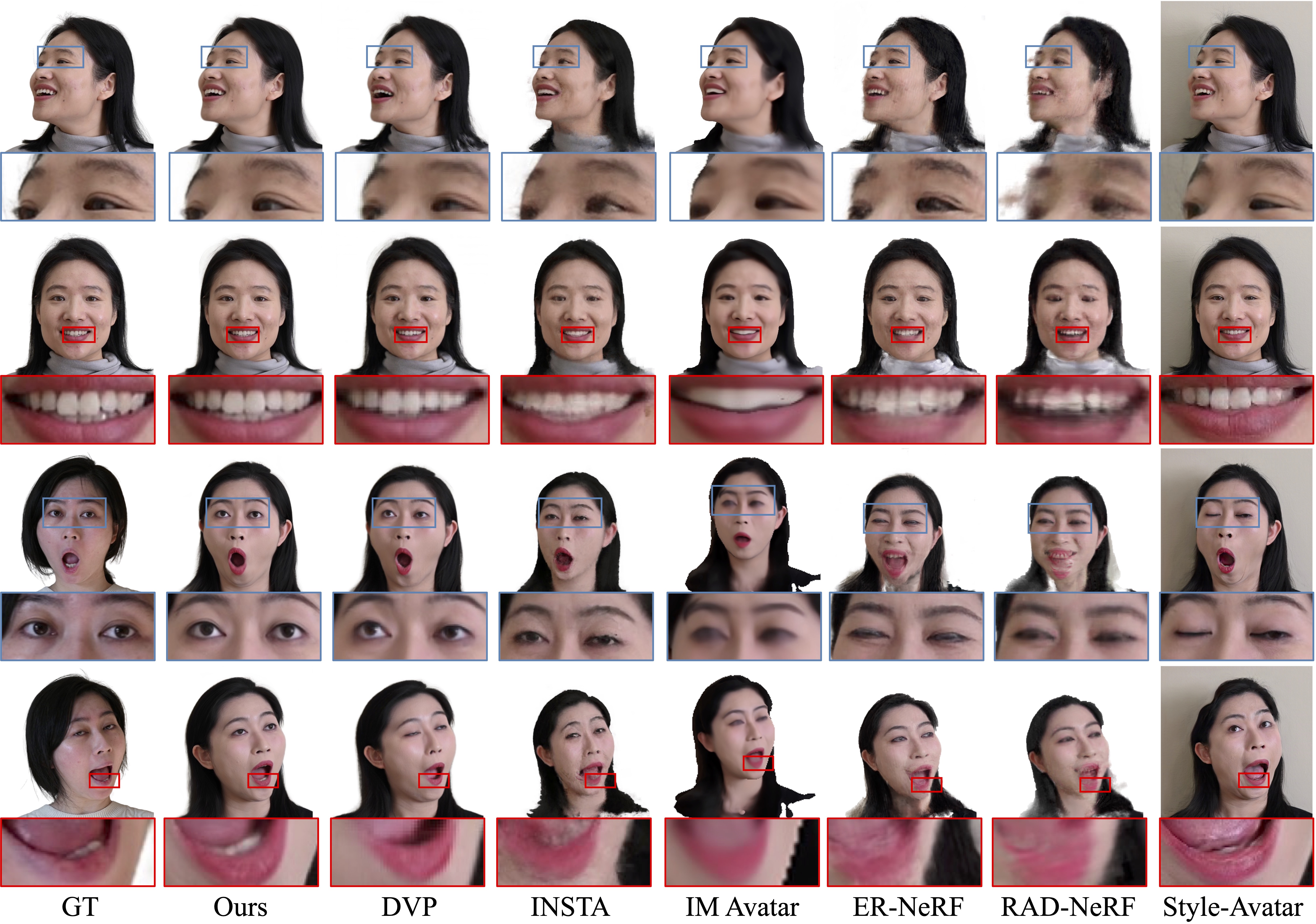}
    \vspace{-0.6cm}
    \caption{
    (1) Qualitative evaluation on the self-reenactment (\textbf{two rows in above}). From left to right: ground truth actor, Ours, Deep Video Portrait~\cite{kim2018deep}, INSTA~\cite{zielonka2023instant}, IM Avatar~\cite{zheng2022avatar}, ER-NeRF~\cite{li2023efficient}, RAD-NeRF~\cite{tang2022real} and Style-Avatar~\cite{wang2023styleavatar} (with background). (2) Qualitative evaluation on the cross-reenactment (\textbf{two rows in below}). We re-animate the portrait with head pose and facial expression from the actors (GT). Since Style-Avatar~\cite{wang2023styleavatar} requires to sample at the scale of $3\times1024\times1024$, which is not available in public self-record videos, so \textbf{we adopt the videos in Dataset A for fairness comparison}. Please refer to \textcolor{red}{Appx. Demo Video} for evaluation. 
    } \label{fig:appendix-4}
\vspace{-0.2cm}
\end{figure*}

\begin{table}[t]
\footnotesize
\begin{center}
\setlength{\tabcolsep}{1.4mm}
{
\vspace{-.1cm}
\begin{tabular}{ccccccccc}
\hlinew{1.15pt}
\multirow{3}{*}{Methods} &F-LMD$\downarrow$ &SD$\downarrow$ &PSNR$\uparrow$&LPIPS$\downarrow$ &MOS$_1$ &MOS$_2$ &MOS$_3$ &MOS$_4$\\
\cline{2-9}
&\multicolumn{4}{c|}{Quantitative Results} &\multicolumn{4}{c}{User Study}\\
\cline{2-9}
&\multicolumn{4}{c|}{Dataset A+B~\cite{Zheng_2023_CVPR,wang2024gaussianhead,zhao2023havatar}}&\multicolumn{4}{c}{Cross-/Self- Reenactment}\\
\hline
IM Avatar & 2.90 & 15.2  &19.39 & 22.0 & 1.54 & 2.20 & 2.71  & 1.90      \\
INSTA &2.81 & 7.73  &25.97  & 9.44 & 2.45 & 2.70 & 1.77 & 2.06     \\
RAD-NeRF & 5.80 & 16.7 & 23.16 & 20.1 & 1.40 & 1.43 & 1.51 & 1.67   \\
ER-NeRF & 3.43 & 13.9 & 21.45 & 28.3  & 1.74 & 1.20  & 1.39  & 1.49 \\
DVP  &2.93 & 5.25 & 25.32  & 8.65 & 3.62 & 3.64 & 3.84  & 3.57   \\
Style-Avatar & 2.64 & 4.78 & 27.83  & 7.53  & \textbf{4.43} & 3.10 & {3.72}  & 3.17    \\
\rowcolor{mygray} Ours &\textbf{2.48} &\textbf{3.50} &\textbf{27.75} & \textbf{5.81} & 4.12 & \textbf{3.95} & \textbf{4.02} & \textbf{4.05} \\
\hlinew{1.15pt}
\end{tabular}}
\vspace{.3cm}
\caption{(1) Left: Quantitative results of IM Avatar~\cite{zheng2022avatar}, INSTA~\cite{zielonka2023instant}, RAD-NeRF~\cite{tang2022real}, ER-NeRF~\cite{li2023efficient}, DVP~\cite{kim2018deep}, Style-Avatar~\cite{wang2023styleavatar} on Dataset A and Dataset B~\cite{Zheng_2023_CVPR,wang2024gaussianhead,zhao2023havatar}. We bold the best. The experiments are performed under self-reconstruction setting. The value of SD and LPIPS are multiplied by $10^{-1}$ and $10^{2}$ respectively. (2) Right: The MOS score for user study. The videos being evaluated include self-/cross- reenactments results}\label{table:appendix-1}
\end{center}
\vspace{-1.cm}
\end{table}

\noindent \textbf{Evaluation Results.} The quantitative results are summarized in Table~\ref{table:appendix-1}. The visualization results are shown in Fig.~\ref{fig:appendix-4}, with the corresponding area are zoomed in.  We still archive the best performance with those baselines. From Fig.~\ref{fig:appendix-4}, it is easy to find the NeRF-based method suffers from image quality reduction (as in the tooth and eyes in ER-NeRF~\cite{li2023efficient}, RAD-NeRF~\cite{tang2022real}, INSTA~\cite{zielonka2023instant}, and IMAvatar~\cite{zheng2022avatar}). While DVP~\cite{kim2018deep} yields good results, it struggles to capture high-frequency details. As the most advanced method, Style-Avatar~\cite{wang2023styleavatar} excels in self-reenactment but meets challenges in maintaining facial structures in cross-reenactment, where the inherent facial structure is compromised due to the illusion of the pre-trained StyleGAN~\cite{karras2019style}, resulting in a decrease in overall realism (it is similar to the issues in HAvatar~\cite{zhao2023havatar}). Our method outperforms baselines in recovering high-frequency details and controlling head pose and expression in both self-/cross- reenactments.

\noindent \textbf{User Study.} We follow the same setting in main paper to perform human evaluation. The results are shown in Table~\ref{table:appendix-1}. Our method scored slightly lower than Style-Avatar in image quality ($4.43$ to $4.12$), but significantly better in video quality ($3.95$ to $3.10$) and motion consistency ($4.02$ to $3.72$), demonstrating fewer artifacts and more consistent performance.

\subsection{Limitations} 
\vspace{-0.1cm}

Tri$^2$-plane achieve promising performance in most cases (shown in figures and demo video). However, it still suffers from several artifacts. There are two most important limitations among them.

\noindent\textbf{Unnatural movements of the torso.} This problems also located in GaussianHead~\cite{wang2024gaussianhead}, HAvatar~\cite{zhao2023havatar}, IMAvatar~\cite{zheng2022avatar} \etc.. Usually, the movement of the torso is not consistent with the movement of the head, and the range of the torso is smaller. We need to introduce additional constraints to control the movement of the torso, as the AD-NeRF~\cite{guo2021ad}, RAD-NeRF~\cite{tang2022real} and ER-NeRF~\cite{li2023efficient}~\etc. Additional face parsing models need to be utilized in which to segment the head and the torso in order to self-supervise the torso. However, most of our actors have Eastern faces and female with long hair, which causes the existing parsing model to be unable to segment different part will (out-of-distribution of parsing model). Following the traditional second-stage (face and torso) decomposition pipeline will result in performance degradation. We hope to solve this problem through global consistency in the future.

\noindent\textbf{Pixel shaking with long hair part.} Alongside torso movement issues, we observe pixel shaking in parts with long hair. Unlike the head, long hair reacts more freely to physical effects, lacking synchronization with head movements. Our current method, which optimizes on a single frame basis, does not account for global consistency, resulting in noticeable instability in hair movement. Moreover, we recognize that existing NeRF-based methods struggle with accurately rendering long hair, an inherent and longstanding issue in the field.

\subsection{Ethical Consideration} 

Our research primarily focuses on simulating high-fidelity facial avatars. However, due to its photo-realistic facial rendering capabilities, there exists a potential for misuse. For example, creating speech videos of public figures portraying events or statements that never occurred. The risk of such abuses is a longstanding concern in the field of AI-synthesized photo-realistic humans, evident in phenomena like deepfake swapping and talking head generation. While it is challenging to completely prevent the misuse of this technology, our paper contributes by providing a technical analysis of facial synthesis. This insight allows users to better understand the field and recognize the limitations of AI synthesis to a certain extent, including aspects like tooth detail and temporal consistency. Furthermore, we advocate for responsible usage practices. These include measures like embedding watermarks in generated videos and employing synthetic face detection technologies~\cite{song2022adaptive,song2022face} for photo-realistic portraits. Such steps are crucial in mitigating the risks associated with this technology.

\appendix

\end{document}